\documentclass[letterpaper]{article} 
\usepackage{aaai2026}  
\usepackage{times}  
\usepackage{helvet}  
\usepackage{courier}  
\usepackage[hyphens]{url}  
\usepackage{graphicx} 
\usepackage{natbib}  
\usepackage{caption} 
\usepackage{algorithm}

\usepackage{algpseudocode}

\usepackage{amsmath}
\usepackage{booktabs}
\usepackage{multirow}
\usepackage{amssymb}
\usepackage{color}
\usepackage{array}
\usepackage{colortbl, xcolor} 
\usepackage{subcaption}

\definecolor{mygray}{RGB}{240, 240, 240} 

\usepackage{newfloat}
\usepackage{listings}
\DeclareCaptionStyle{ruled}{labelfont=normalfont,labelsep=colon,strut=off} 
\floatstyle{ruled}
\newfloat{listing}{tb}{lst}{}
\floatname{listing}{Listing}
\newcommand{\mllm}{FaceShield}

\newcommand{\cls}{coarse-grained classification}
\newcommand{\attack}{fine-grained classification}
\newcommand{\reason}{reasoning}
\newcommand{\loc}{attack localization}
\newcommand{\visionencoder}{Spoof-Aware Vision Perception}
\def\datasetpre{FaceShield-pre10K}
\def\datasetsft{FaceShield-sft45K}

\def\modelname{\mllm\includegraphics[height=2.8ex]{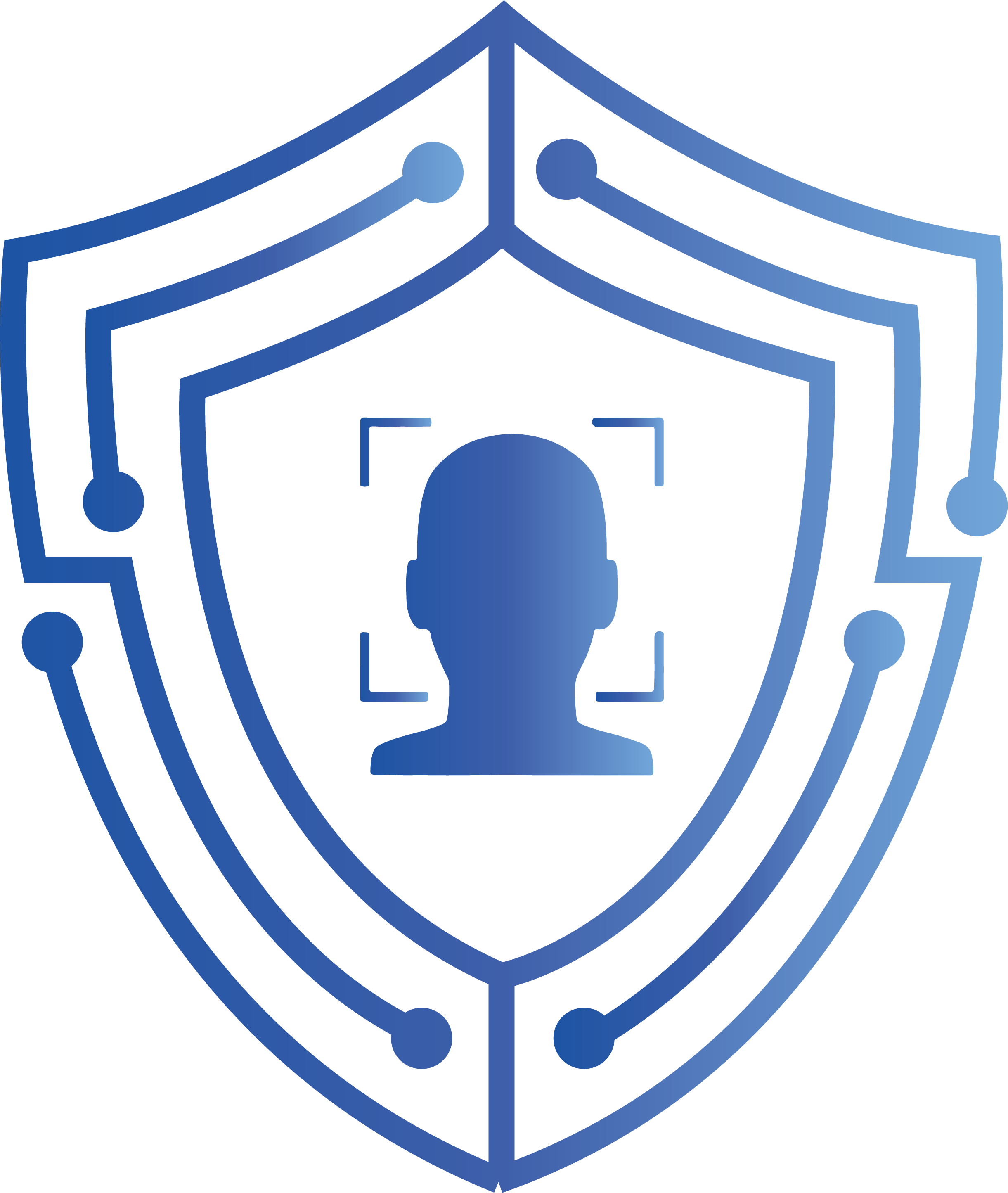}: Explainable Face Anti-Spoofing with \\Multimodal Large Language Models}

\title{\modelname}

\author{
    Hongyang Wang\textsuperscript{\rm 1,2}\equalcontrib,
    Yichen Shi\textsuperscript{\rm 3,4}\equalcontrib,
    Zhuofu Tao\textsuperscript{\rm 4,5},
    Yuhao Gao\textsuperscript{\rm 1,2},
    Liepiao Zhang\textsuperscript{\rm 6},
    Xun Lin\textsuperscript{\rm 7}\\
    Jun Feng\textsuperscript{\rm 1,2},
    Xiaochen Yuan\textsuperscript{\rm 10},
    Zitong Yu\textsuperscript{\rm 7,8,9}\thanks{Corresponding author.},
    Xiaochun Cao\textsuperscript{\rm 11}
}
\affiliations{
    \textsuperscript{\rm 1}Shijiazhuang Tiedao University
    \textsuperscript{\rm 2}Shijiazhuang Key Laboratory of Artifical Intelligence\\
    \textsuperscript{\rm 3}Shanghai Jiao Tong University
    \textsuperscript{\rm 4}Eastern Institute of Technology
    \textsuperscript{\rm 5}University of California, Los Angeles\\
    \textsuperscript{\rm 6}GRGBanking
    \textsuperscript{\rm 7}Great Bay University
    \textsuperscript{\rm 10}Macao Polytechnic University\\
    
    \textsuperscript{\rm 11}Shenzhen Campus of Sun Yat-sen University
    \textsuperscript{\rm 9}Dongguan Key Laboratory for Intelligence and Information Technology\\

    \textsuperscript{\rm 8}Guangdong Provincial Key Laboratory of Intelligent Information Processing \& Shenzhen Key Laboratory of Media Security, Shenzhen University\\

}

\usepackage{bibentry}

\begin{document}

\twocolumn[{
    \renewcommand\twocolumn[1][]{#1}
    \maketitle
    \vspace{-1.5mm}
    \begin{center}
        \centering
        \includegraphics[width=0.98\textwidth]{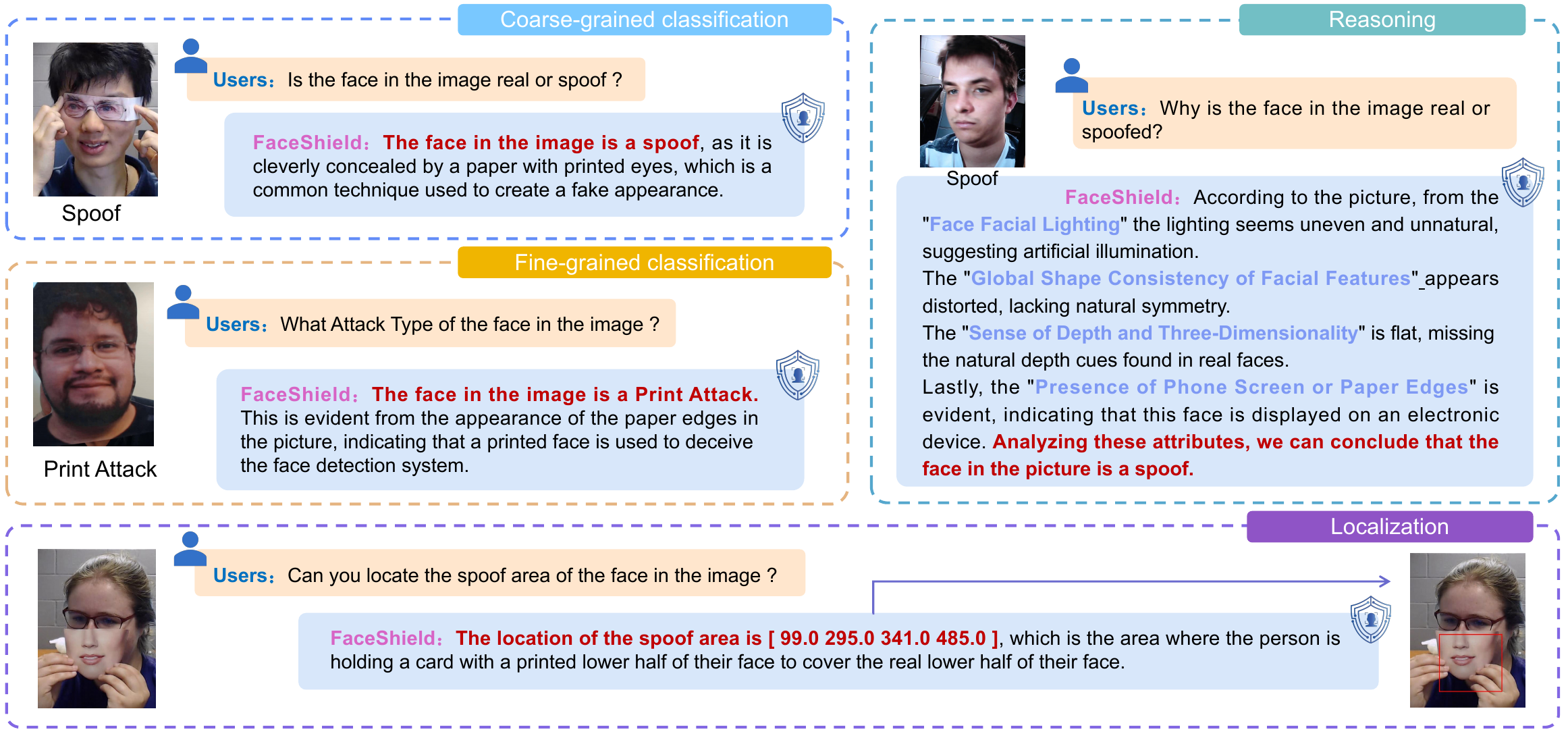}
        \captionof{figure}{FaceShield Multi-task Response Demonstration. This figure shows the model's performance on four tasks: coarse-grained classification (real vs. spoofed faces), fine-grained classification (specific attack types like print attacks), reasoning (explaining spoofing using features such as lighting and symmetry), and localization (detecting spoofed regions). It highlights FaceShield's ability to handle diverse, complex questions accurately.} 
        \label{fig:teaser} 
    \end{center}
}]

\begin{abstract}
Face anti-spoofing (FAS) is crucial for protecting facial recognition systems from presentation attacks. Previous methods approached this task as a classification problem, lacking interpretability and reasoning behind the predicted results. Recently, multimodal large language models (MLLMs) have shown strong capabilities in perception, reasoning, and decision-making in visual tasks. However, there is currently no universal and comprehensive MLLM and dataset specifically designed for FAS task. To address this gap, we propose FaceShield, a MLLM for FAS, along with the corresponding pre-training and supervised fine-tuning (SFT) datasets, FaceShield-pre10K and FaceShield-sft45K. FaceShield is capable of determining the authenticity of faces, identifying types of spoofing attacks, providing reasoning for its judgments, and detecting attack areas. Specifically, we employ spoof-aware vision perception (SAVP) that incorporates both the original image and auxiliary information based on prior knowledge. We then use an prompt-guided vision token masking (PVTM) strategy to random mask vision tokens, thereby improving the model's generalization ability. We conducted extensive experiments on three benchmark datasets, demonstrating that FaceShield significantly outperforms previous deep learning models and general MLLMs on four FAS tasks, i.e., coarse-grained classification, fine-grained classification, reasoning, and attack localization. Our instruction datasets, protocols, and codes will be released at \url{https://github.com/Why0912/FaceShield}.
\end{abstract}



\section{Introduction}

Face anti-spoofing (FAS) is essential in facial recognition systems, ensuring that presentation attacks (PAs), such as print, replay, and 3D wearable masks, are effectively prevented. It has attracted considerable interest in industry and academia in the past decade. 

\begin{figure}
    \centering
    \includegraphics[width=1\linewidth]{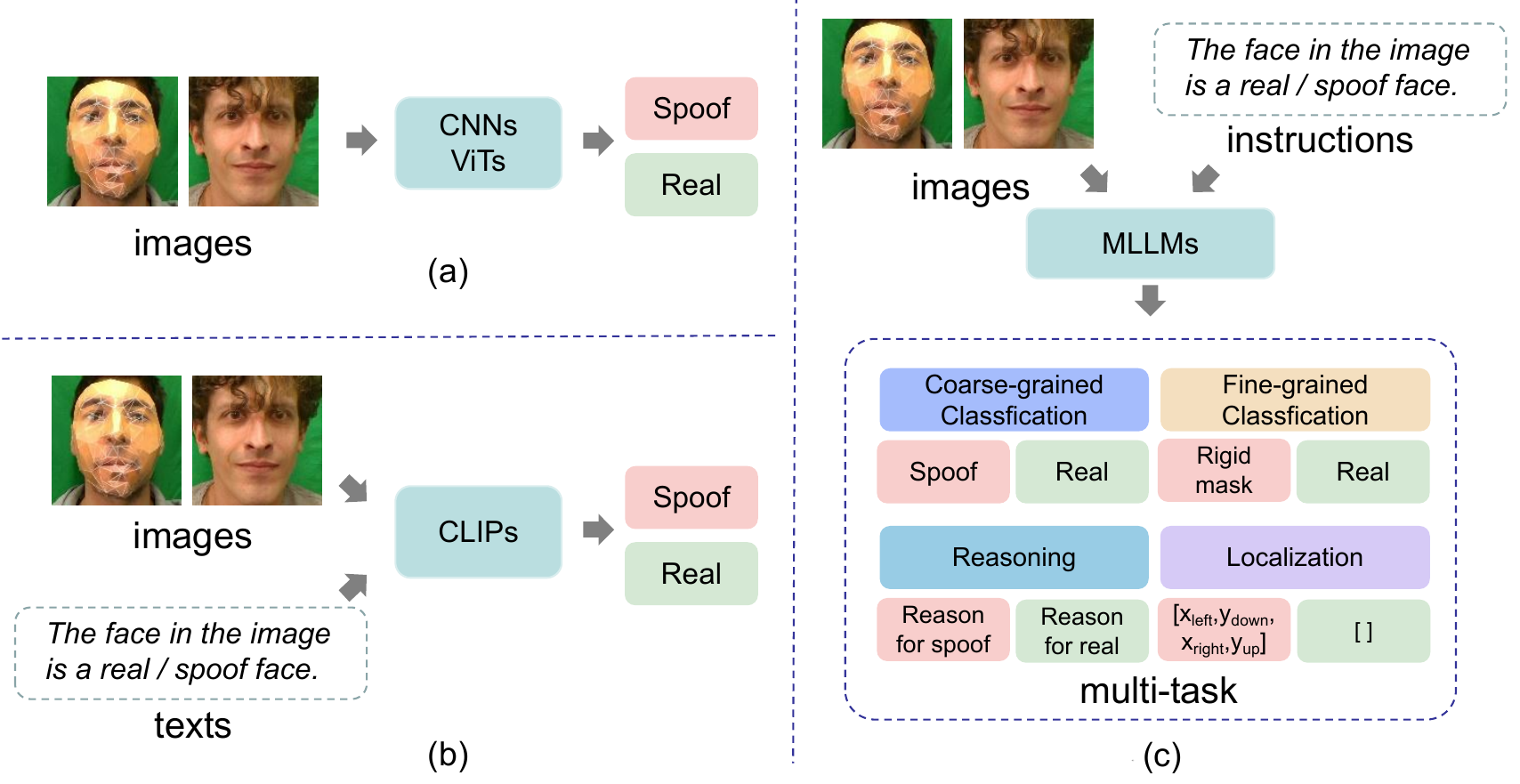}
    \caption{Pipelines of different FAS methods (a) traditional deep learning models, (b) multimodal models, and (c) MLLM }
    \label{fig:existing fas}
\end{figure}

Existing deep learning FAS models can be categorized into two types: vision-based methods and vision-language-based methods. As shown in Fig.~\ref{fig:existing fas}(a), vision-based methods rely solely on image data (e.g., RGB, Depth, Infrared(IR)) and binary labels to train CNNs~\cite{yu2020searching,yu2021dual,wang2025pnss} or ViTs~\cite{9484333,yu2023rethinkingvisiontransformermasked,cai2025rehearsal} for FAS. While they can achieve satisfactory results against known attack types and environments, these methods are prone to overfitting on spurious correlations and lack strong extrapolation capabilities. As illustrated in Fig.~\ref{fig:existing fas}(b), Vision-language-based methods do not use binary labels but instead train CLIPs with image-text pairs~\cite{Srivatsan_2023_ICCV,liu2024fmclip,mu2024tegdgtextuallyguideddomain,lin2025reliable}. The text labels in these methods provide more domain-agnostic information, enhancing models’ generalization capability. Although these FAS models demonstrate some recognition capabilities, they still face challenges such as limited generalization ability, poor interpretability, and a lack of capability for fine-grained localization of attack regions.

Recently, MLLMs have shown remarkable capabilities across various visual tasks~\cite{ye2025cat}, such as remote sensing~\cite{zhang2024earthgpt,kuckreja2023geochat,2402.02544}, medical imaging~\cite{li2023llavamed,Sun2024STLLaVAMedSL}, and deepfake detection ~\cite{xu2024fakeshieldexplainableimageforgery,huang2024ffaa}. By leveraging the general capabilities of language foundation models alongside the visual information extracted by vision towers, these specialized MLLMs integrate perception, reasoning, and decision-making within a single model. Regarding the FAS task, SHIELD~\cite{shi2025shield} conducted extensive evaluations on existing general-purpose MLLMs, revealing that their performance on FAS tasks still has room for improvement. \cite{zhang2025interpretable} introduced a model capable of performing classification and description attack type. However, the model is limited in its ability to handle more nuanced tasks, such as identifying specific attack types, reasoning, and localizing spoofed areas. These limitations underscore the broader challenges in training FAS MLLMs, including: (1) a lack of pretraining and SFT datasets specific to FAS tasks, (2) the need to extend traditional FAS tasks to fully exploit MLLM capabilities, and (3) the difficulty for general-purpose vision towers to capture the subtle distinctions between real faces and PAs, unlike with natural images.

Motivated by the above discussion, in this paper, we expand the traditional FAS task to include four sub-tasks (see Fig.~\ref{fig:teaser} for examples): \cls, \attack, \reason, and \loc. We then introduce FaceShield, an MLLM specifically designed for these tasks. As can be seen from Fig.~\ref{fig:existing fas}(c), we propose a pretraining and SFT dataset generation pipeline. This pipeline constructs two multimodal FAS instruction datasets containing 50k dialogues for \mllm\ training. To the best of our knowledge, \mllm\ is the first FAS MLLM, equipped with multiple detection capabilities. Additionally, \datasetpre\ and \datasetsft\ are the first high-quality datasets that can be used to train a FAS-specific MLLM. Our main contributions include:
\begin{itemize} 
    \item We develop a novel data generation pipeline that utilizes a MLLM and predefined prompts, and construct two multimodal FAS instruction datasets (i.e., \datasetpre\ and \datasetsft) with 12 attack types. To our best knowledge, these are the first multitask instruction datasets for the FAS community.

    \item We propose \mllm, the first multitask MLLM for FAS that is capable of \cls, \attack, \reason, and \loc. \mllm\  utilizes the \visionencoder\ (SAVP) and Prompt-guided Visual Token Masking (PVTM)  strategies to enhance the discrimination of confusing attack areas.

    \item Extensive experiments demonstrate that \mllm\ significantly outperforms previous specialized FAS models and general MLLMs across multiple datasets in various FAS evaluation tasks.
    
\end{itemize}

\label{sec:intro}

\begin{figure*}[t]
    \centering
    \includegraphics[width=1\linewidth]{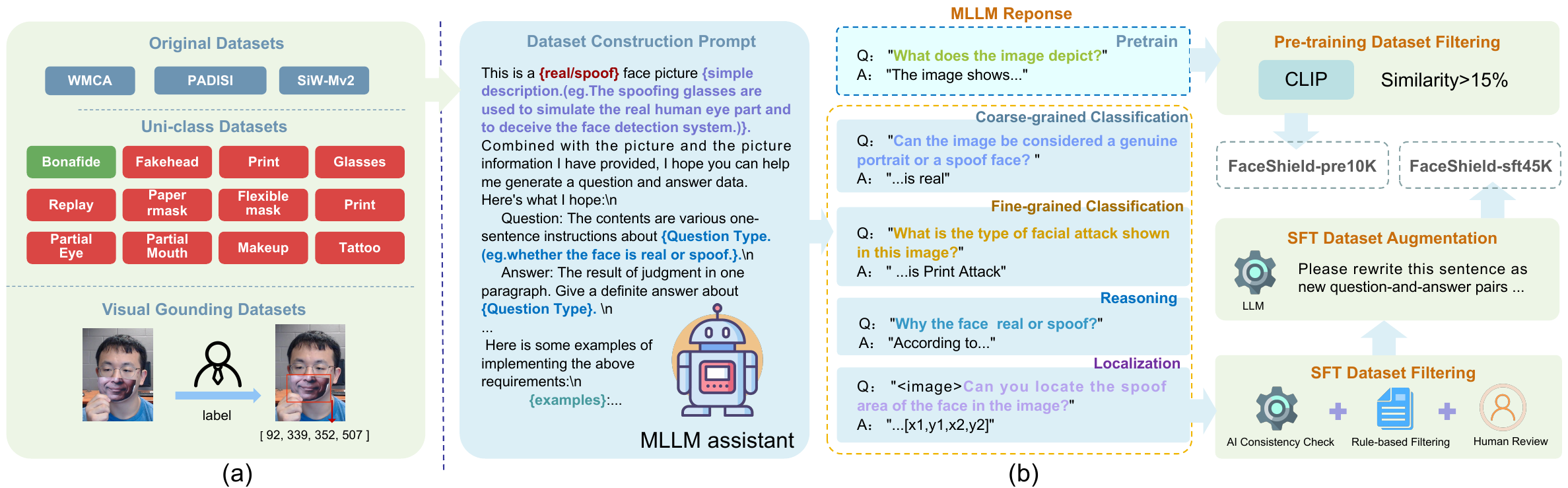}
    \caption{Construction pipeline of our proposed  instruction datasets (i.e., \datasetpre\ and \datasetsft).The initial datasets (WMCA, PADISI, SiW-Mv2) are combined to form a uni-class dataset covering 12 spoofing types, with selected images annotated for visual grounding. Using MLLM with structured prompts, we generate two datasets: a pretraining dataset and an SFT dataset divided into four tasks (coarse-grained classification, fine-grained classification, reasoning, and localization). The pretraining data is filtered by CLIP for similarity, producing the FaceShield-pre10k dataset. SFT data undergoes multi-level filtering (LLM-based, keywords, and human reviews), followed by augmentation, resulting in the FaceShield-sft45k dataset. Additional details can be found in the appendix.}
    \label{fig:dataset}
\end{figure*}

\section{Related Work}
\noindent\textbf{Face Anti-Spoofing. }
Early FAS methods primarily used CNNs~\cite{yu2020searching} and ViTs~\cite{9484333}, incorporating auxiliary cues such as reflection~\cite{icassp22multiple}, depth~\cite{cvpr18aux}, and rPPG~\cite{yu2019remote} for live/spoof classification. While some fuse multi-modal inputs (e.g., RGB, IR, depth)\cite{yu2024rethinking}, they often suffer performance drops in unseen domains due to domain shifts\cite{tpami23survey}. To enhance generalization, domain-generalized FAS methods~\cite{tpami23survey} employ techniques such as adversarial training~\cite{ijcv23adversarial}, feature disentanglement~\cite{cvpr22ssan}, one-class learning~\cite{cvpr24oneclass}, meta-learning~\cite{qin2021meta}, data augmentation~\cite{ijcv24augmentation}, and data synthesis~\cite{aaai22generation} to model domain shifts and separate domain-invariant content from domain-specific styles. Additionally, domain or gradient alignment~\cite{cvpr24gradient} aids in learning generalized decision boundaries for more robust and domain-invariant spoof detection.

Recently, vision-language models (VLMs), especially CLIP~\cite{clip}, have been explored for FAS~\cite{flip,liu2024fmclip}. These models use textual descriptions of live and spoof faces to guide classification, offering stronger generalization by leveraging the powerful visual representations learned during pre-training~\cite{liu2024fmclip}.

However, most FAS methods lack interpretability, particularly in providing language-based explanations. A recent work~\cite{zhang2025interpretable} collected 12 datasets and used instruction tuning to train an MLLM for attack classification, but it struggles to integrate classification, localization, and reasoning tasks for real-world deployment. Additionally, existing FAS datasets~\cite{eccv25tffas,cvpr24cfpl} lack reasoning annotations, hindering the development of more comprehensive and interpretable models.

\noindent\textbf{Multimodal Large Language Models. }
MLLMs such as GPT-4V~\cite{gpt4}, LLaVA~\cite{liu2023llava}, and Bunny~\cite{he2024bunny} have shown strong performance across general domains~\cite{li2023seedbench2benchmarkingmultimodallarge,huang2024aesbenchexpertbenchmarkmultimodal}. Task-specific MLLMs have also been developed for various vision tasks, including remote sensing~\cite{zhang2024earthgpt,kuckreja2023geochat}, medical imaging~\cite{li2023llavamed,Sun2024STLLaVAMedSL}, aesthetic assessment~\cite{huang2024aesexpert}, deepfake detection~\cite{xu2024fakeshieldexplainableimageforgery,huang2024ffaa}, and visual grounding~\cite{wei2023lenna,peng2023kosmos}. Unlike traditional models, MLLMs unify multiple tasks within a single framework and generalize well to unseen domains. For example, FakeShield~\cite{xu2024fakeshieldexplainableimageforgery} introduces an explainable multi-task model for image forgery detection and localization by leveraging visual-textual clues at both global and pixel levels.

However, current MLLMs lack specific knowledge about face attacks, leading to limited performance on FAS tasks. To address this, we introduce dedicated FAS datasets that enrich MLLMs with domain-specific knowledge. The proposed SAVP module enhances spoof discrimination by guiding the model with semantic attack priors, while PVTM improves feature generalization across varied domains.

\section{\datasetpre\ and \datasetsft\ }

\subsection{Dataset Collection}

\begin{figure*}[t]
    \centering
    \includegraphics[width=1\linewidth]{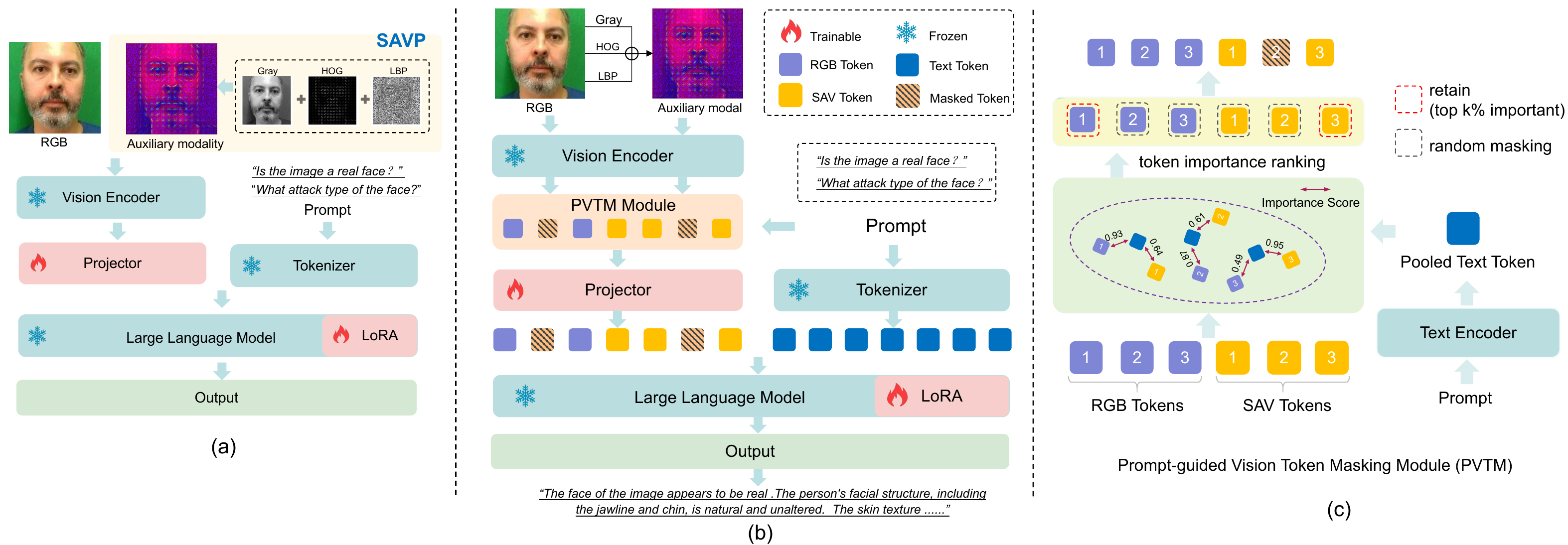}
    \caption{Proposed model architectures. (a) Proposed model with Spoof-Aware Vision Perception (SAVP). (b) Proposed model with SAVP and Prompt-Guided Vision Token Masking (PVTM). (c) Details about PVTM.}
    \label{fig:network}
\end{figure*}

Fig.~\ref{fig:dataset}(a) illustrates the annotation process for existing FAS datasets. Based on the class types from WMCA (W)~\cite{8714076}, PADSIS (P)~\cite{Rostami_2021_ICCV}, and SiW-Mv2 (S)~\cite{xiaoguo2022MDFAS}, we unify the annotation categories into 12 types: Bonafide, Fakehead, Print, Glasses, Replay, Paper mask, Flexible mask, Rigid mask, Partial Eye, Partial Mouth, Makeup, and Tattoo. We re-annotate all images at both image- and region-levels, resulting in 12,091 images with class labels and 3,139 images with bounding box annotations. The detail information is shown in Table~\ref{tab:dataset_qa}.

\subsection{Instruction Construction}

As shown in Fig.~\ref{fig:dataset}(b), we construct two instruction datasets using Bunny-Llama-3-8B-V~\cite{he2024bunny}. A system prompt containing class labels is used to guide the MLLM assistant in generating question-answer (QA) pairs, based on the task type and few-shot examples.

For the pretraining dataset \datasetpre, we generate image descriptions only, without task instructions. Pairs with a CLIP~\cite{radford2021learning} similarity score below 15\% are filtered out to ensure quality.

\begin{algorithm}[H]
\caption{Construction of FaceShield-sft45K Dataset}
\label{alg:FaceShield-sft45K}
\begin{algorithmic}[1]
\small
\Require
    Image dataset $\mathcal{D}_{img}$ from WMCA, SiW-Mv2, PADISI; \\
    Multimodal LLM (MLLM); Task set $\mathcal{T}$.
\Ensure
    Processed QA dataset $\mathcal{D}_{QA}$.

\State Initialize $\mathcal{D}_{QA} \gets \emptyset$
\For {image $I \in \mathcal{D}_{img}$}
    \For {task $t \in \mathcal{T}$}
        \State Generate task-specific prompt $\mathcal{P}_t$ using Ground Truth (GT), requirements and examples
        \State Generate QA pair $(q, a)$ using MLLM: $(q, a) = \text{MLLM}(\mathcal{P}_t, I)$
        \If {QA pair $(q, a)$ satisfies validation criteria}
            \State Add $(q, a)$ to $\mathcal{D}_{QA}$
        \EndIf
    \EndFor
\EndFor

\State Filter low-quality QA pairs in $\mathcal{D}_{QA}$:
\State \hspace{1em} Remove pairs with low semantic consistency or invalid content
\State \hspace{1em} Perform human review to refine remaining pairs

\State Augment $\mathcal{D}_{QA}$ by generating rephrased QA pairs:
\For {QA pair $(q, a) \in \mathcal{D}_{QA}$}
    \State Generate additional pairs $(q', a')$ using rephrasing strategies
    \State Add $(q', a')$ to $\mathcal{D}_{QA}$
\EndFor

\State \Return $\mathcal{D}_{QA}$
\end{algorithmic}
\end{algorithm}

For the instruction-tuning dataset \datasetsft, as shown in Alg.~\ref{alg:FaceShield-sft45K}, 
 MLLM-generated QA pairs are filtered using both manual and keyword-based strategies. The resulting high-quality seed set is then diversified using LLaMA3~\cite{dubey2024llama} to enhance linguistic richness and dialogue ability.

\begin{table}[ht]
\centering
\caption{Datasets and QA statistics}
\resizebox{0.47\textwidth}{!}{
\begin{tabular}{l l l l}
\toprule
Datasets & Attack Types & Annotations & QA\_Count \\
\midrule
FaceShield-pre10K & - & Image-text pairs & 9297 \\
FaceShield-sft45K & Unified-attack(11 types) & QA + bbox & 45662 \\
\bottomrule
\end{tabular}
}
\label{tab:dataset_qa}
\end{table}

The dataset covers four tasks: (1) \textbf{\textit{Coarse-grained classification}}, which predicts whether a face is real or spoofed; (2) \textbf{\textit{Fine-grained classification}}, which identifies specific PA types beyond binary classification; (3) \textbf{\textit{Reasoning}}, which provides explanations and justifications before making a judgment; (4) \textbf{\textit{Attack localization}}, which outputs coordinates of attack regions if spoofing is detected.

\label{sec:dateset}

\section{\mllm}
Our goal is to train a FAS task-specific MLLM with two main objectives: 1) Enhance the visual encoder's ability to extract features from real faces and presentation attacks, and 2) Utilize the extensive knowledge stored in the LLM to improve the model's generalization capabilities when facing unknown domains.A naive training approach involves direct pre-training and SFT using RGB images and constructed QA data. However, the high similarity between real faces and PAs in RGB appearance poses significant challenges to this method. As shown in Fig.~\ref{fig:network}(a), Spoof-Aware Vision Perception (SAVP) combines images preprocessed based on prior knowledge, by extracting predefined local descriptor operators~\cite{yu2024rethinking}, with the original RGB image. Our complete model framework is shown in Fig.~\ref{fig:network}(b), RGB images and the extracted local descriptor images are fed into the vision encoder to extract vision token $V_{RGB}$ and $V_{SAV}$, respectively. These features are then processed through the Prompt-Guided Vision Token Masking (PVTM) module, which extracts highly generalizable vision tokens. These tokens are sent to a projector to align with text prompt token $P$ to produce $V_{align}$, which is then fed into the language foundation model for inference result $\mathcal{Y}$ as follows:
\begin{equation}
V_{align} = \text{Projection}(V_{RGB}, V_{SAV})
\end{equation}
\begin{equation}
\mathcal{Y} = \text{MLLM}(V_{align}, P)
\end{equation}

\subsection{Spoof-Aware Vision Perception}
Bonafide faces and PAs lack distinct discriminative features in RGB-based appearance space, whereas local descriptors~\cite{yu2020searching,yu2024rethinking,xie2024fusionmamba} extracted through image preprocessing can enhance their subtle live/spoof clues. As shown in Fig.~\ref{fig:network}(a), we extract features from the original images using Local Binary Pattern (LBP) \cite{pietikainen2010local}, Gray, and Histogram of Oriented Gradients (HOG) \cite{dalal2005histograms}, and concatenate them. LBP and Gray-specific computations are as follows:
\begin{equation}
\begin{aligned}
LBP &= \sum_{i=0}^{P-1} s(g_i - g_c) \cdot 2^i, \\
s(x) &= \begin{cases}
1 & \text{if } x \geq 0 , \\
0 & \text{otherwise} ,
\end{cases}
\end{aligned}
\end{equation}
where $g_c$ is the pixel value of the central pixel in the considered neighborhood, and $g_i$ represents the pixel values of the $P$ surrounding pixels.
\begin{equation}
\text{Gray}(I) = 0.299 \cdot R + 0.587 \cdot G + 0.114 \cdot B,
\label{eq:img2gray}
\end{equation}
where $R$, $G$, and $B$ are the red, green, and blue intensity values of the pixel, respectively.

The HOG calculates the gradient magnitude and direction at each pixel using edge detection operators and then divides the image into small, overlapping cells. Within each cell, gradients are binned according to their direction into histograms. Histograms from the cells within each block are concatenated and normalized based on the block's overall gradient energy. The final HOG descriptor is formed by the vector of these normalized histograms from all blocks.

We perform the above three steps of feature extraction on the original image, then concatenate these as three channels to form a complete image. This composite image is then fed into the vision encoder to extract features as complementary information. The spoof-aware vision token $V_{SAV}$ and final vision input $V$ for \mllm\ as follows: 
\begin{equation}
V_{SAV} = \text{Encoder}(\text{Concat}[LBP, Gray, HOG])
\label{eq:img2gray}
\end{equation}
\begin{equation}
V = \text{Concat}[V_{RGB}, V_{SAV}])
\label{eq:img2gray}
\end{equation}

\subsection{Prompt-Guided Vision Token Masking}
To further enhance the alignment between visual features and text prompts, and alleviate overfitting on spurious correlations, we leverage text prompts to guide token selection after the visual encoder. As shown in Fig.~\ref{fig:network}(c), the text tokens extracted from the prompt are pooled~\cite{song2024moresimpleeffectivetoken}, and then their similarity with all visual tokens is calculated. We assume that visual tokens with higher similarity are more relevant to subsequent tasks and even for the same image, the important visual tokens may not be consistent across different tasks. The calculation of similarity between visual tokens $V_i$ and text prompt tokens $P$ is as follows:
\begin{equation}
Sim(V_i, P) = \frac{V_i \cdot P}{\|V_i\| \|P\|}
\end{equation}

Subsequently, we apply a softmax function to the similarities between all $V_i$ and $P$, using the resulting values $S^i_{rank}$ as an importance metric for each visual token. We then rank all tokens based on this importance calculated as follows:
\begin{equation}
S^i_{rank}(V_i, P) = \frac{e^{S(V_i, P)}}{\sum_j e^{S(V_j, P)}}
\end{equation}

Afterward, we retain the top $k\%$ of vision tokens in importance. The remaining tokens are then randomly masked with a probability of $p\%$, reducing the influence of less important tokens while keeping acceptable information loss for the final decision-making process.

\subsection{Training Details}
\label{sec:method}
We adopt a two-stage training strategy. In the first stage (pretraining), we perform continual pretraining to align visual embeddings from a pretrained vision encoder with text embeddings using \datasetpre. In the second stage (supervised fine-tuning, SFT), we apply visual instruction tuning on \datasetsft\ to fully exploit the MLLM's capabilities across domain-specific multimodal tasks.

For LLM adaptation, we apply LoRA~\cite{hu2021loralowrankadaptationlarge} with a rank of 128 and a scaling factor of 256 for each transformer block. Cross-entropy loss is used for next-token prediction in both stages.
During pretraining, we update only the vision projector and the PVTM module for one epoch. In the SFT stage, we fine-tune the LoRA layers in the LLM along with the vision projector.

\section{Experiments and Results}

\subsection{Protocols and Evaluation Metrics}
We use 10\% of each source dataset in \datasetsft\ to construct three test subsets: W, S, and P. For \cls\ we perform both intra- and cross-dataset evaluations. For \attack, \reason\ and \loc, we conduct intra-dataset evaluation only. In intra-dataset testing, models are trained on all source data and evaluated on the combined test sets (W\&S\&P). For cross-dataset testing, two datasets are used for training (including pretraining and SFT), and the remaining one for testing (e.g., training on W and S, testing on P).For \cls, \attack, and \reason, we report Half Total Error Rate (HTER)~\cite{yu2022deep} and Accuracy (ACC). For \reason, we further evaluate reasoning quality using BLEU~\cite{papineni2002bleu}, ROUGE-L~\cite{lin2004rouge}, and METEOR~\cite{banerjee2005meteor}. For \loc, we report AP@40 and AP@50.

\subsection{Implementation Details}

We use Siglip~\cite{zhai2023sigmoid} as the visual encoder and Phi-3~\cite{abdin2024phi} as the language foundation model. PVTM retains the top 10\% of the most important tokens and randomly masks 5\% of the tokens in the remaining 90\%. Adam optimizer is used in the pretrain stage with a learning rate of \(5 \times 10^{-4}\). As for SFT stage, we decrease the learning rate to \(2 \times 10^{-4}\). All experiments are conducted on a single NVIDIA A100 GPU. Each experiment is repeated 10 times on the model, and the final results are reported as the mean ± standard deviation.

\subsection{Comparison with Existing Methods}

\subsubsection{Coarse-Grained Classification Task}

For the coarse-grained classification task, we compare FaceShield with state-of-the-art FAS methods ~\cite{he2016deep,Wang_2022_CVPR, zhou2022learning,zhou2023instance} and open-source MLLMs~\cite{liu2023llava,Qwen-VL,zhu2023minigpt,he2024bunny}.

\begin{table}
\centering
\caption{Intra-dataset results on coarse-grained classification.}
\label{tab:Intra-domain—cls}
\scalebox{0.85}{  
\begin{tabular}{lcc}
\toprule[1pt]
Method                 & \multicolumn{1}{l}{ACC(\%) $\uparrow$ } & \multicolumn{1}{l}{HTER(\%) $\downarrow$ } \\ \hline
\rowcolor{mygray} 
\multicolumn{3}{c}{\textbf{Traditional}}   \\
ResNet \cite{he2016deep}           &97.55   & 2.32 \\
PatchNet \cite{Wang_2022_CVPR}            & 98.22    & 1.78      \\
CoOp~\cite{zhou2022learning} & 98.73  & 1.27  \\ \hline
\rowcolor{mygray} 
\multicolumn{3}{c}{\textbf{MLLM}}    \\
LLaVA \cite{liu2023llava}                & 65.54    & 27.76     \\
Qwen-VL \cite{Qwen-VL}        & 51.94    & 38.70     \\
Minigpt4 \cite{zhu2023minigpt}             & 26.86    & 65.50     \\
Bunny \cite{he2024bunny}               & 81.20    & 17.87     \\
Bunny (fine-tuned)~\cite{he2024bunny}      & 98.23    & 1.52     \\
\textbf{FaceShield (Ours)}                & \textbf{99.41 ± 0.06} & \textbf{0.53 ± 0.06} \\ 
\bottomrule[1pt]
\end{tabular}}
\end{table}

\begin{table}
\centering
\caption{Cross-dataset results on coarse-grained classification. $W$, $S$, and $P$ denote WMCA, SiW-Mv2, and PADISI, respectively.}
\label{tab:Inter-domain}
\scalebox{0.93}{\begin{tabular}{lcc}
\toprule[1pt]
Method                 & \multicolumn{1}{l}{ACC(\%) $\uparrow$ } & \multicolumn{1}{l}{HTER(\%) $\downarrow$ } \\ \hline
\rowcolor{mygray} 
\multicolumn{3}{c}{\textbf{W \& S \text{\textrightarrow} P}}  \\
ResNet \cite{he2016deep}                & 46.12                        & 50.00                        \\
PatchNet \cite{Wang_2022_CVPR}          & 77.18                        & 22.87                        \\
IADG \cite{zhou2023instance}            & 72.96                        & 27.01                        \\
FAS-AUG \cite{ijcv24augmentation}        & 91.7                         & 7.3                          \\  
\textbf{FaceShield (Ours)}              & \textbf{93.17 ± 0.22}        & \textbf{6.37 ± 0.21}         \\ \hline

\rowcolor{mygray} 
\multicolumn{3}{c}{\textbf{W \& P \text{\textrightarrow} S}}           \\
ResNet \cite{he2016deep}                & 53.36                        & 49.16                        \\
PatchNet \cite{Wang_2022_CVPR}          & 56.16                        & 45.37                        \\
IADG \cite{zhou2023instance}            & 57.20                        & 42.81                        \\
FAS-AUG \cite{ijcv24augmentation}          & 88.2                         & 11.7                         \\ 
\textbf{FaceShield (Ours)}              & \textbf{89.93 ± 0.15}        & \textbf{10.3 ± 0.14}         \\ \hline

\rowcolor{mygray} 
\multicolumn{3}{c}{\textbf{S \& P \text{\textrightarrow} W}}        \\
ResNet \cite{he2016deep}                & 74.01                        & 29.75                        \\
PatchNet \cite{Wang_2022_CVPR}          & 78.15                        & 41.50                        \\
IADG \cite{zhou2023instance}            & 78.55                        & 26.27                        \\
FAS-AUG \cite{ijcv24augmentation}         & 87.9                         & 13.1                         \\ 
\textbf{FaceShield (Ours)}              & \textbf{92.56 ± 0.08}        & \textbf{5.71 ± 0.08}         \\ \bottomrule[1pt]
\end{tabular}}
\end{table}

\begin{table}[H]
\centering
\caption{Cross-dataset result on CASIA-MFSD and Replay-Attack.}
\label{tab:cross-to-CI}
\resizebox{\linewidth}{!}{
\begin{tabular}{lcc}
\toprule[1pt]
Test Dataset & ACC(\%) $\uparrow$ & HTER(\%) $\downarrow$ \\
\midrule
S \& P \& W $\rightarrow$ CASIA-MFSD     & 90.59 & 6.37 \\
S \& P \& W $\rightarrow$ Replay-Attack  & 82.42 & 20.07 \\
\bottomrule[1pt]
\end{tabular}
}
\end{table}

\begin{table}[H]
\centering
\caption{Results of fine-grained classification task.}
\label{tab:Attack}
\resizebox{0.95\linewidth}{!}{
\begin{tabular}{lc}
\toprule[1pt]
Method & ACC(\%) $\uparrow$ \\ \midrule
LLaVA \cite{liu2023llava}         & 16.39 \\
Qwen-VL \cite{Qwen-VL}            & 16.55 \\
Minigpt4 \cite{zhu2023minigpt}    & 19.51 \\
Bunny \cite{he2024bunny}          & 27.03 \\
Bunny (Fine-tuned) \cite{he2024bunny} & 94.43 \\
\textbf{FaceShield (Ours)}        & \textbf{95.81 ± 0.11} \\
\bottomrule[1pt]
\end{tabular}
}
\end{table}

\begin{table*}[h]
\centering
\caption{Results of the reasoning task with metrics BLEU, ROUGE-L, METEOR, ACC, and HTER.}
\label{tab:Reasoning}
\resizebox{\textwidth}{!}{%
\begin{tabular}{lcccccccc}
\toprule[1pt]
Method         & BLEU-1 (\%) $\uparrow$ & BLEU-2 (\%) $\uparrow$ & BLEU-3 (\%) $\uparrow$ & BLEU-4 (\%) $\uparrow$ & ROUGE-L (\%) $\uparrow$ & METEOR (\%) $\uparrow$ & ACC (\%) $\uparrow$ & HTER (\%) $\downarrow$ \\ \hline
LLaVA \cite{liu2023llava}          & 45.05 & 31.75 & 23.42 & 17.80 & 30.51 & 25.52 & 37.84 & 50.11 \\
Minigpt4 \cite{zhu2023minigpt}      & 17.85 & 7.85  & 3.53  & 1.94  & 27.54 & 21.88 & 33.86 & 50.00 \\
Qwen-VL \cite{Qwen-VL}   & 20.92 & 14.53 & 11.01 & 8.77  & 21.45 & 12.49 & 47.64 & 41.49 \\
Bunny \cite{he2024bunny}          & 33.64 & 27.12 & 22.65 & 19.33 & 36.74 & 19.12 & 50.68 & 39.73 \\
Bunny(fine-tuned) \cite{he2024bunny}     & 89.57 & 86.96 & 84.91 & 81.29 & 80.15 & 51.64 & 98.56 & 1.16  \\
\textbf{FaceShield (Ours)}        & \textbf{90.89 ± 0.14
} & \textbf{88.02 ± 0.15} & \textbf{85.75 ± 0.17
} & \textbf{83.98 ± 0.19
} & \textbf{82.98 ± 0.20
} & \textbf{53.10 ± 0.16} & \textbf{99.29 ± 0.04} & \textbf{0.57 ± 0.04}  \\ \bottomrule[1pt]
\end{tabular}}

\end{table*}

\begin{table}[t]
\centering
\caption{Results of the attack localization task with metrics AP@40 and AP@50.}
\label{tab:Location}
 \scalebox{0.8}{\begin{tabular}{lcc}
\toprule[1pt]
Method         & AP@40 (\%) $\uparrow$ & AP@50 (\%) $\uparrow$ \\ \hline
Qwen-VL \cite{Qwen-VL}   & 2.07 & 1.49 \\
Lenna \cite{wei2023lenna}          & 37.77 & 35.41 \\
Sphinx \cite{lin2023sphinxjointmixingweights}        & 47.86 & 46.30 \\
Bunny \cite{he2024bunny}          & 73.50 & 71.65 \\
Bunny(fine-tuned) \cite{he2024bunny}  & 92.30 & 89.71 \\
\textbf{FaceShield (Ours)}          & \textbf{97.78 ± 0.21} & \textbf{95.60 ± 0.19} \\ \bottomrule[1pt]
\end{tabular}}
\end{table}

\begin{table}[t]
\caption{Ablation study on pretraining w/ or w/o FaceShield-pre10K dataset.}
\centering
\resizebox{\columnwidth}{!}{%
\begin{tabular}{cccc}
\toprule[1pt]
\multicolumn{1}{c}{\parbox{2cm}{\centering \datasetpre}} &
\multicolumn{1}{c}{\parbox{4cm}{\centering Fine-grained\\Classification}} &
\multicolumn{2}{c}{Reasoning} \\
\cline{3-4}
& ACC (\%) $\uparrow$ & ACC (\%) $\uparrow$ & HTER (\%) $\downarrow$ \\ \hline
$\times$     & 94.78 $\pm$ 0.19 & 98.83 $\pm$ 0.06 & 0.94 $\pm$ 0.05 \\
$\checkmark$ & \textbf{95.81 $\pm$ 0.11} & \textbf{99.29 $\pm$ 0.04} & \textbf{0.57 $\pm$ 0.04} \\
\bottomrule[1pt]
\end{tabular}%
}
\label{tab:ab_nopre_text}
\end{table}

 \paragraph{Intra-dataset Testing.}
Table~\ref{tab:Intra-domain—cls} demonstrates that Our FaceShield significantly outperforms three representative traditional FAS methods~\cite{he2016deep,Wang_2022_CVPR, zhou2022learning}. Moreover, our performance greatly exceeds the zero-shot capabilities of general MLLM. We also fine-tune the open-source MLLM (i.e., Bunny \cite{he2024bunny}), selecting RGB images and language data from the dataset to conduct experiments on Bunny. We find that FaceShield also surpasses the well-tuned MLLM (Bunny) with 1\% HTER decrease. 

\paragraph{Cross-dataset Testing.}
Table~\ref{tab:Inter-domain} shows the performance of \mllm\ in cross-domain scenarios, where we trained on two out of three selected datasets and tested on one. \mllm\ demonstrates performance far exceeding traditional FAS models in cross-domain scenarios. Under the S\&P\text{\textrightarrow}W protocol, it achieves the HTER of 5.72\%, showcasing \mllm's strong generalization capabilities compared to traditional methods. Further results in Table~\ref{tab:cross-to-CI} confirm its robustness across unseen datasets like CASIA-MFSD and Replay-Attack.

\begin{table*}[t]
\centering
\caption{Ablation results across different tasks.}
\label{tab:ab_savp_PVTM}
\resizebox{\textwidth}{!}{%
\begin{tabular}{ccccccccc} 
\toprule[1pt]
\multicolumn{1}{c}{SAVP} & PVTM &
\multicolumn{2}{c}{Coarse-grained classification} &
\multicolumn{1}{c}{Fine-grained classification} &
\multicolumn{2}{c}{Reasoning} &
\multicolumn{2}{c}{Attack localization} \\
\cline{3-9}
 &  & ACC (\%) $\uparrow$ & HTER (\%) $\downarrow$ & ACC (\%) $\uparrow$ &
 ACC (\%) $\uparrow$ & HTER (\%) $\downarrow$ & AP@40 (\%) $\uparrow$ & AP@50 (\%) $\uparrow$ \\ \hline
$\times$     & $\times$     & 98.23 & 1.52 & 94.43 & 98.56 & 1.16 & 92.30 & 89.71 \\
$\times$     & $\checkmark$ & 98.32 & 1.83 & 95.06 & 98.70 & 1.04 & 92.21 & 90.82 \\
$\checkmark$ & $\times$     & 98.73 & 1.06 & 94.59 & \textbf{99.41} & \textbf{0.48} & 97.09 & 95.21 \\
$\checkmark$ & $\checkmark$ &
\textbf{99.41 $\pm$ 0.06} & \textbf{0.53 $\pm$ 0.06} &
\textbf{95.81 $\pm$ 0.11} & 99.29 $\pm$ 0.04 & 0.57 $\pm$ 0.04 &
\textbf{97.78 $\pm$ 0.21} & \textbf{95.6 $\pm$ 0.19} \\
\bottomrule[1pt]
\end{tabular}%
}
\end{table*}

\subsubsection{Fine-Grained Classification Task}
Table~\ref{tab:Attack} shows the results under the fine-grained classification task. For open-source MLLMs, we incorporated 12 types of attacks into the prompt, allowing it to respond with the correct type. For the fine-tuned MLLM and our \mllm, we selected keywords from the responses for evaluation. It is evident that supervised fine-tuning can significantly improve the model's performance, with \mllm~achieving the best results. 

\subsubsection{Reasoning Task}
We also explore the models' reasoning capacity and Table~\ref{tab:Reasoning} displays the results for the \reason\ task. General MLLMs perform poorly in both classification and reasoning. \mllm\ significantly outperforms general MLLMs in both the reasoning process and judgment, and also exceeds open-source MLLMs (e.g., Bunny) fine-tuned on our dataset. \mllm\ not only provides accurate results but also delivers detailed and correct reasoning, effectively enhancing the explainability of FAS methods.

\subsubsection{Attack Localization Task}
It is vital to locate the spoof regions for explainable FAS, and Table~\ref{tab:Location} presents the results for the attack localization task. Due to the scarcity of attack localization annotations in general pre-trained datasets, general MLLMs perform poorly on this task. In contrast, \mllm\ shows excellent results, achieving over 95\% for both AP@40 and AP@50. It accurately locates attack areas, providing new insights into attack region detection for FAS tasks.

\begin{figure}[t]
    \centering
    \includegraphics[width=1\linewidth]{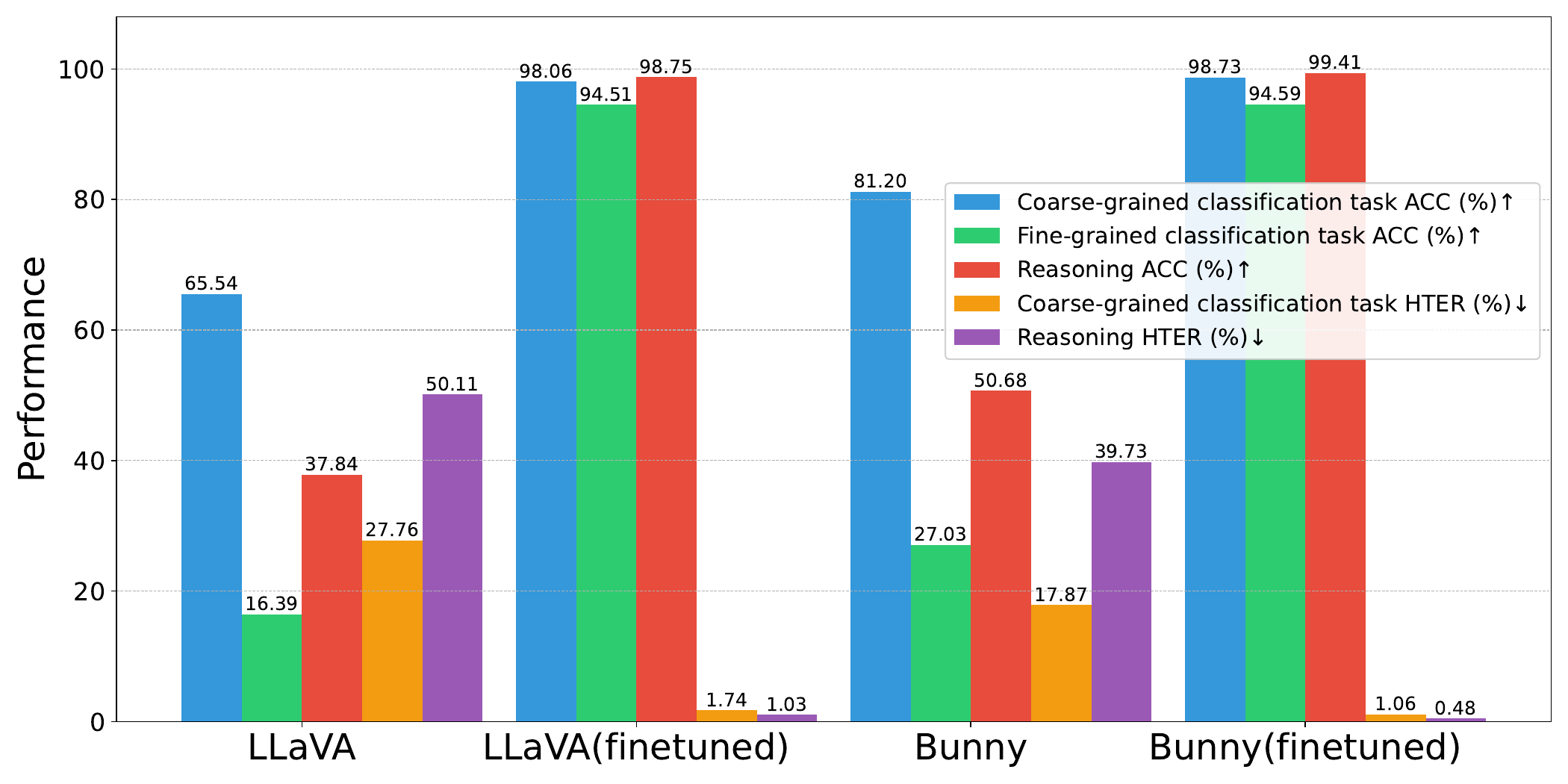}
    \caption{Comparison of performance after fine-tuning using our proposed dataset on LLaVA and Bunny models.}
    \label{fig:ab_dataset}
\end{figure}

\begin{figure}[t]
    \centering
    \includegraphics[width=1\linewidth]{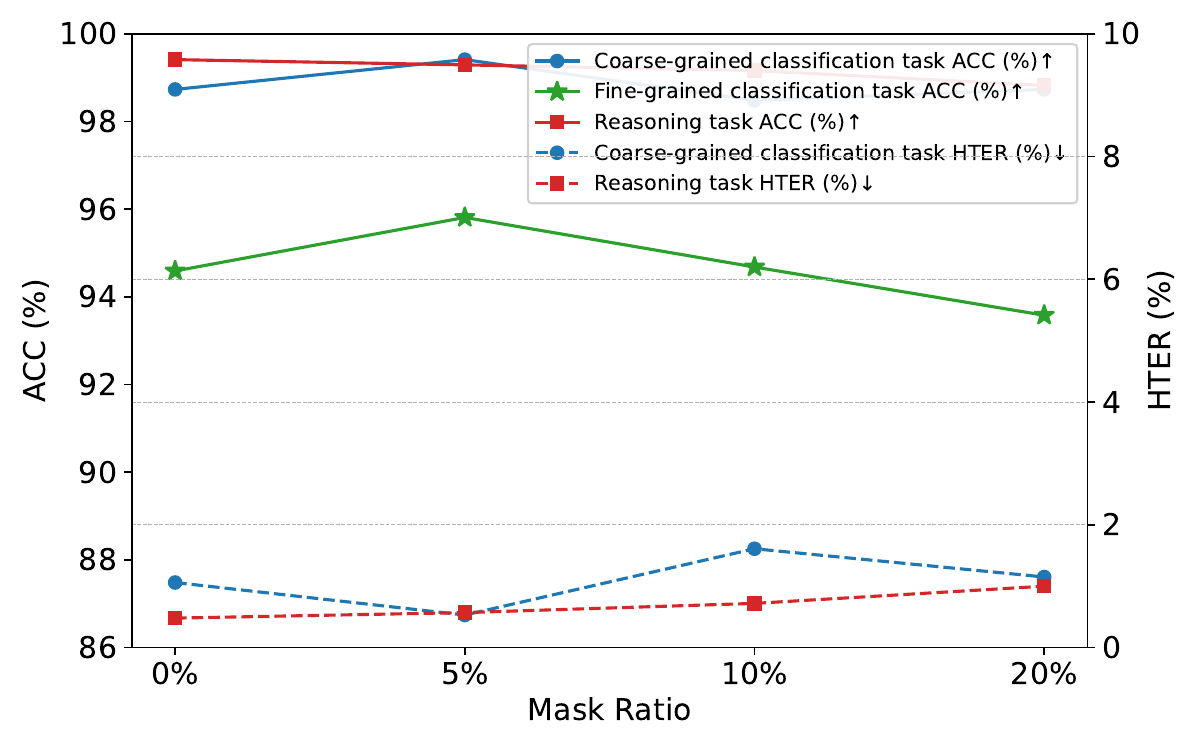}
    \caption{Ablation of visual token masking ratio $p$ in PVTM on  three tasks (i.e., coarse- \& fine-grained classification, and reasoning)}
    \label{fig:ablation_mask}
\end{figure}

\begin{figure}[t]
    \centering
    \includegraphics[width=1\linewidth]{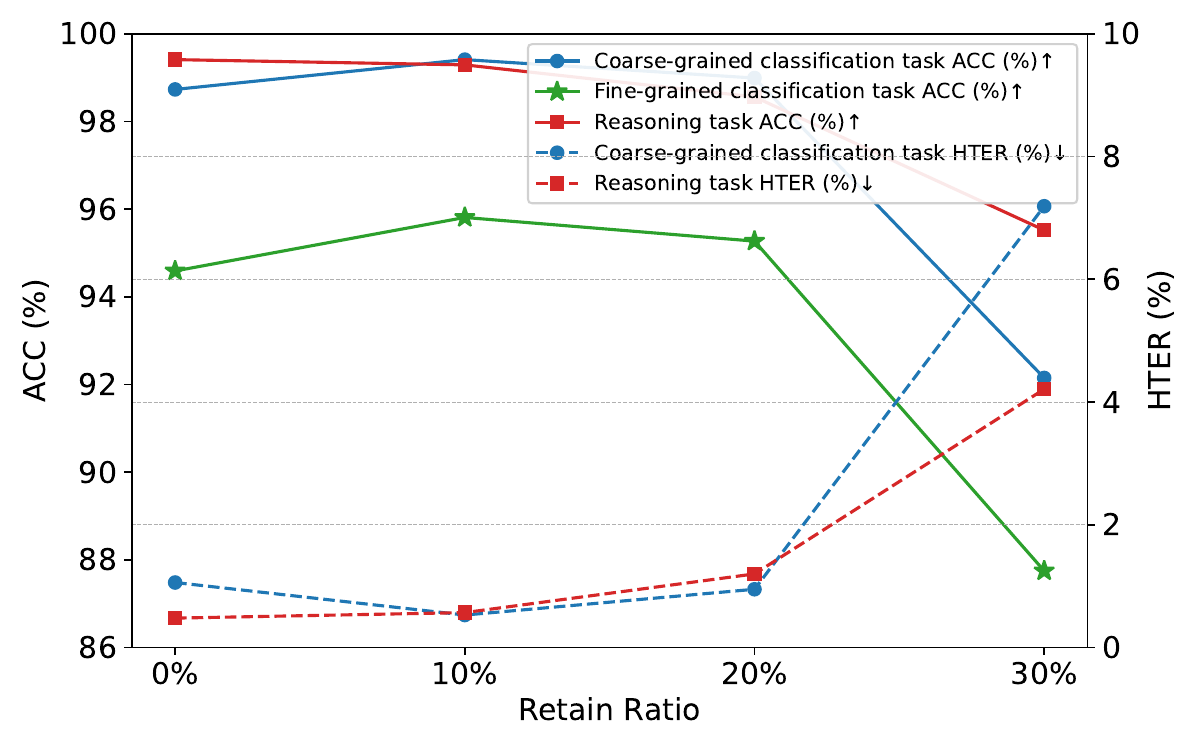}
    \caption{Ablation of visual token retain ratio $k$ in PVTM on three tasks (i.e., coarse- \& fine-grained classification, and reasoning)}
    \label{fig:ablation_retain}
\end{figure}

\subsection{Ablation Study}

\textbf{Effectiveness of the proposed datasets. }
We conduct pre-training and SFT with our proposed FaceShield-pre10K and FaceShield-sft45K on LLaVA \cite{liu2023llava} and Bunny \cite{he2024bunny} to evaluate the efficacy and generalization of our constructed datasets. As shown in Table \ref{tab:ab_nopre_text}, pretraining with the FaceShield-pre10K dataset significantly improves performance, with fine-grained classification accuracy and reasoning accuracy increasing, while HTER is reduced. This demonstrates that pretraining with FaceShield-pre10K enhances the model's capabilities in FAS-related tasks.

Additionally, the results in Fig. \ref{fig:ab_dataset} show that fine-tuning on our dataset further boosts performance across three tasks for both LLaVA and Bunny. This validates the effectiveness of our dataset in enriching MLLMs with FAS-related knowledge and improving their overall performance in FAS tasks, supporting the efficacy of our advanced dataset construction pipeline.

\noindent\textbf{Effectiveness of SAVP. }
Results from the first two rows in Table \ref{tab:ab_savp_PVTM} show that leveraging local descriptors as complementary visual inputs significantly improves performance across four tasks, particularly in the attack localization task, where AP@40 and AP@50 increased by 5.6\% and 5.94\%, respectively. It indicates that prior knowledge-based auxiliary information significantly enhances the model’s ability to distinguish easily confusable facial images. The local live/spoof details within the auxiliary data proves especially valuable for fine-grained attack region detection tasks.

\noindent\textbf{Effectiveness of PVTM. }
It can be seen from the last two rows of Table \ref{tab:ab_savp_PVTM} that the proposed PVTM provides reasonable improvements for FaceShield across three (coarse- and fine-grained classification, and \loc) tasks. It indicates that masking less important tokens helps prevent the model from task-unrelated noises and spurious correlations. However, PVTM leads to a slight performance decrease on the reasoning task. This might be because masking partial visual tokens may compromise overall image perception and lose information for reasoning.

We further study PVTM with different visual token masking ratios $p$, as shown in Fig.\ref{fig:ablation_mask}. Through experiments on three (coarse- and fine-grained classification, and \reason) tasks, we find that masking 5\% of the tokens strikes an optimal balance between reducing spurious correlations and preserving essential information. However, no improvement is found when masking more tokens due to severe information loss. Additionally, we investigate varying proportions of visual token retain ratio $k$, as illustrated in Fig.\ref{fig:ablation_retain}. We preserve 0\%, 10\%, 20\%, and 30\% of the tokens, respectively, and then randomly mask $p$=5\% of the remaining tokens. The results show that retaining $k$=10\% visual tokens with strong importance achieves optimal performance. However, the more tokens preserved, the poorer the model performs, suggesting that as the importance of tokens decreases, the likelihood of spurious correlations increases.

\section{Visualization and Analysis}

\begin{figure}[H]
    \centering
    \includegraphics[width=1\linewidth]{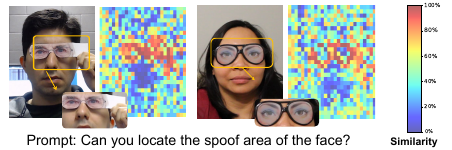}
    \caption{Importance visualization of SAV tokens for attack localization task.}
    \label{fig:bonafide_fakehead_mask}
\end{figure}

Fig.~\ref{fig:bonafide_fakehead_mask} illustrates the visualization obtained after applying SAVP to the \loc\ task, highlighting the effectiveness of SAV tokens. Compared to RGB tokens, which may suffer from dispersed attention, SAV tokens, once applied, can accurately focus on spoofed regions (e.g., eyeglass areas, hand-held masks). These tokens show strong alignment with the ground-truth annotations, producing more focused and interpretable activations around spoof artifacts. Additionally, visual tokens in the attack regions exhibit higher similarity scores with the task prompt. With PVTM applied, the model is further guided to focus on these deception-relevant areas, emphasizing the critical roles of both SAV tokens and PVTM in improving localization performance.

\section{Conclusion}
\label{sec:Conclusion}
In this paper, we expand the FAS task into four sub-tasks: \cls, \attack, \reason, and \loc, and propose the \mllm, a specialized MLLM tailored for these FAS tasks.
Considering the specific training data requirements of MLLM, we establish a pipeline for constructing datasets tailored for FAS task pre-training and supervised fine-tuning, resulting in the creation of the \datasetpre\ and \datasetsft\ datasets. This paper represents a preliminary exploration of MLLM for FAS task, and future works will focus on incorporating multiple visual modalities and refining the granularity of attack region localization.


\bibliography{aaai2026}

@String(IJCV = {Int. J. Comput. Vis.})

@String(CVPR= {IEEE Conf. Comput. Vis. Pattern Recog.})

@String(ICCV= {Int. Conf. Comput. Vis.})

@String(ECCV= {Eur. Conf. Comput. Vis.})

@String(ICASSP=	{ICASSP})

@String(IJCAI = {IJCAI})

@String(AAAI = {AAAI})

@String(IJCV  = {IJCV})

@String(CVPR  = {CVPR})

@String(ICCV  = {ICCV})

@String(ECCV  = {ECCV})

@article{ye2025cat,
  title={Cat+: Investigating and enhancing audio-visual understanding in large language models},
  author={Ye, Qilang and Yu, Zitong and Shao, Rui and Cui, Yawen and Kang, Xiangui and Liu, Xin and Torr, Philip and Cao, Xiaochun},
  journal={IEEE TPAMI},
  year={2025}
}

@article{shi2025shield,
  title={Shield: An evaluation benchmark for face spoofing and forgery detection with multimodal large language models},
  author={Shi, Yichen and Gao, Yuhao and Lai, Yingxin and Wang, Hongyang and Feng, Jun and He, Lei and Wan, Jun and Chen, Changsheng and Yu, Zitong and Cao, Xiaochun},
  journal={Visual Intelligence},
  volume={3},
  number={1},
  pages={9},
  year={2025},
  publisher={Springer}
}

@inproceedings{yu2020searching,
    title={Searching Central Difference Convolutional Networks for Face Anti-Spoofing},
    author={Yu, Zitong and Zhao, Chenxu and Wang, Zezheng and Qin, Yunxiao and Su, Zhuo and Li, Xiaobai and Zhou, Feng and Zhao, Guoying},
    booktitle= {CVPR},
    year = {2020}
}

@misc{2402.02544,
Author = {Dilxat Muhtar and Zhenshi Li and Feng Gu and Xueliang Zhang and Pengfeng Xiao},
Title = {LHRS-Bot: Empowering Remote Sensing with VGI-Enhanced Large Multimodal Language Model},
Year = {2024},
Eprint = {arXiv:2402.02544},
archivePrefix={arXiv},
}

@article{zhang2024earthgpt,
  title={Earthgpt: A universal multi-modal large language model for multi-sensor image comprehension in remote sensing domain},
  author={Zhang, Wei and Cai, Miaoxin and Zhang, Tong and Zhuang, Yin and Mao, Xuerui},
  journal={IEEE Transactions on Geoscience and Remote Sensing},
  year={2024},
  publisher={IEEE}
}

@article{kuckreja2023geochat,
          title={GeoChat: Grounded Large Vision-Language Model for Remote Sensing},
          author={Kuckreja, Kartik and Danish, Muhammad S. and Naseer, Muzammal and Das, Abhijit and Khan, Salman and Khan, Fahad S.},
          journal={The IEEE/CVF Conference on Computer Vision and Pattern Recognition},
          year={2024}
  }

@article{li2023llavamed,
  title={Llava-med: Training a large language-and-vision assistant for biomedicine in one day},
  author={Li, Chunyuan and Wong, Cliff and Zhang, Sheng and Usuyama, Naoto and Liu, Haotian and Yang, Jianwei and Naumann, Tristan and Poon, Hoifung and Gao, Jianfeng},
  journal={arXiv preprint arXiv:2306.00890},
  year={2023}
}

@inproceedings{Sun2024STLLaVAMedSL,
  title={STLLaVA-Med: Self-Training Large Language and Vision Assistant for Medical},
  author={Guohao Sun and Can Qin and Huazhu Fu and Linwei Wang and Zhiqiang Tao},
  booktitle = {EMNLP},
  year={2024},
}

@article{huang2024ffaa,
         title={FFAA: Multimodal Large Language Model based Explainable Open-World Face Forgery Analysis Assistant},
         author={Huang, Zhengchao and Xia, Bin and Lin, Zicheng and Mou, Zhun and Yang, Wenming},
         journal={arXiv preprint arXiv:2408.10072},
         year={2024}
}

@misc{xu2024fakeshieldexplainableimageforgery,
      author = {Zhipei Xu and Xuanyu Zhang and Runyi Li and Zecheng Tang and Qing Huang and Jian Zhang},
      title = {FakeShield: Explainable Image Forgery Detection and Localization via Multi-modal Large Language Models},
      year = {2024},
      note = {arXiv preprint arXiv:2410.02761},
      url = {https://arxiv.org/abs/2410.02761},
}

@InProceedings{Wang_2022_CVPR,
    author    = {Wang, Chien-Yi and Lu, Yu-Ding and Yang, Shang-Ta and Lai, Shang-Hong},
    title     = {PatchNet: A Simple Face Anti-Spoofing Framework via Fine-Grained Patch Recognition},
    booktitle = {Proceedings of the IEEE/CVF Conference on Computer Vision and Pattern Recognition (CVPR)},
    month     = {June},
    year      = {2022},
    pages     = {20281-20290}
}

@inproceedings{yu2021dual,
    title={Dual-Cross Central Difference Network for Face Anti-Spoofing},
    author={Yu, Zitong and Qin, Yunxiao and Zhao, Hengshuang and Li, Xiaobai and Zhao, Guoying},
    booktitle= {IJCAI},
    year = {2021}
}

@INPROCEEDINGS{9484333,
  author={George, Anjith and Marcel, Sébastien},
  booktitle={2021 IEEE International Joint Conference on Biometrics (IJCB)}, 
  title={On the Effectiveness of Vision Transformers for Zero-shot Face Anti-Spoofing}, 
  year={2021},
  volume={},
  number={},
  pages={1-8},
  doi={10.1109/IJCB52358.2021.9484333}}

@misc{yu2023rethinkingvisiontransformermasked,
      title={Rethinking Vision Transformer and Masked Autoencoder in Multimodal Face Anti-Spoofing}, 
      author={Zitong Yu and Rizhao Cai and Yawen Cui and Xin Liu and Yongjian Hu and Alex Kot},
      year={2023},
      eprint={2302.05744},
      archivePrefix={arXiv},
      primaryClass={cs.CV},
      url={https://arxiv.org/abs/2302.05744}, 
}

@InProceedings{Srivatsan_2023_ICCV,
    author    = {Srivatsan, Koushik and Naseer, Muzammal and Nandakumar, Karthik},
    title     = {FLIP: Cross-domain Face Anti-spoofing with Language Guidance},
    booktitle = {Proceedings of the IEEE/CVF International Conference on Computer Vision (ICCV)},
    month     = {October},
    year      = {2023},
    pages     = {19685-19696}
}

@inproceedings{
liu2024fmclip,
title={{FM}-{CLIP}: Flexible Modal {CLIP} for Face Anti-Spoofing},
author={Ajian Liu and Ma Hui and Junze Zheng and Haocheng Yuan and Xiaoyuan Yu and Yanyan Liang and Sergio Escalera and Jun Wan and Zhen Lei},
booktitle={ACM Multimedia 2024},
year={2024},
url={https://openreview.net/forum?id=Gl3a5nusJP}
}

@misc{mu2024tegdgtextuallyguideddomain,
      title={TeG-DG: Textually Guided Domain Generalization for Face Anti-Spoofing}, 
      author={Lianrui Mu and Jianhong Bai and Xiaoxuan He and Jiangnan Ye and Xiaoyu Liang and Yuchen Yang and Jiedong Zhuang and Haoji Hu},
      year={2024},
      eprint={2311.18420},
      archivePrefix={arXiv},
      primaryClass={cs.CV},
      url={https://arxiv.org/abs/2311.18420}, 
}

@ARTICLE{8714076,
  author={George, Anjith and Mostaani, Zohreh and Geissenbuhler, David and Nikisins, Olegs and Anjos, André and Marcel, Sébastien},
  journal={IEEE Transactions on Information Forensics and Security}, 
  title={Biometric Face Presentation Attack Detection With Multi-Channel Convolutional Neural Network}, 
  year={2020},
  volume={15},
  number={},
  pages={42-55},
  doi={10.1109/TIFS.2019.2916652}}

@InProceedings{Rostami_2021_ICCV,
    author    = {Rostami, Mohammad and Spinoulas, Leonidas and Hussein, Mohamed and Mathai, Joe and Abd-Almageed, Wael},
    title     = {Detection and Continual Learning of Novel Face Presentation Attacks},
    booktitle = {Proceedings of the IEEE/CVF International Conference on Computer Vision (ICCV)},
    month     = {October},
    year      = {2021},
    pages     = {14851-14860}
}

@inproceedings{xiaoguo2022MDFAS,
    title={Multi-domain Learning for Updating Face Anti-spoofing Models},
    author={Guo, Xiao and Liu, Yaojie and Jain, Anil and Liu, Xiaoming},
    booktitle={ECCV},
    year={2022}
}

@misc{liu2023llava,
      title={Visual Instruction Tuning}, 
      author={Liu, Haotian and Li, Chunyuan and Wu, Qingyang and Lee, Yong Jae},
      publisher={NeurIPS},
      year={2023},
}

@article{Qwen-VL,
  title={Qwen-VL: A Versatile Vision-Language Model for Understanding, Localization, Text Reading, and Beyond},
  author={Bai, Jinze and Bai, Shuai and Yang, Shusheng and Wang, Shijie and Tan, Sinan and Wang, Peng and Lin, Junyang and Zhou, Chang and Zhou, Jingren},
  journal={arXiv preprint arXiv:2308.12966},
  year={2023}
}

@article{zhu2023minigpt,
  title={MiniGPT-4: Enhancing Vision-Language Understanding with Advanced Large Language Models},
  author={Zhu, Deyao and Chen, Jun and Shen, Xiaoqian and Li, Xiang and Elhoseiny, Mohamed},
  journal={arXiv preprint arXiv:2304.10592},
  year={2023}
}

@article{he2024bunny,
      title={Efficient Multimodal Learning from Data-centric Perspective}, 
      author={He, Muyang and Liu, Yexin and Wu, Boya and Yuan, Jianhao and Wang, Yueze and Huang, Tiejun and Zhao, Bo},
      journal={arXiv preprint arXiv:2402.11530},
      year={2024}
}

@article{wei2023lenna,
  title={Lenna: Language enhanced reasoning detection assistant},
  author={Wei, Fei and Zhang, Xinyu and Zhang, Ailing and Zhang, Bo and Chu, Xiangxiang},
  journal={arXiv preprint arXiv:2312.02433},
  year={2023}
}

@misc{lin2023sphinxjointmixingweights,
      title={SPHINX: The Joint Mixing of Weights, Tasks, and Visual Embeddings for Multi-modal Large Language Models}, 
      author={Ziyi Lin and Chris Liu and Renrui Zhang and Peng Gao and Longtian Qiu and Han Xiao and Han Qiu and Chen Lin and Wenqi Shao and Keqin Chen and Jiaming Han and Siyuan Huang and Yichi Zhang and Xuming He and Hongsheng Li and Yu Qiao},
      year={2023},
      eprint={2311.07575},
      archivePrefix={arXiv},
      primaryClass={cs.CV},
      url={https://arxiv.org/abs/2311.07575}, 
}

@inproceedings{he2016deep,
  title={Deep residual learning for image recognition},
  author={He, Kaiming and Zhang, Xiangyu and Ren, Shaoqing and Sun, Jian},
  booktitle={Proceedings of the IEEE conference on computer vision and pattern recognition},
  pages={770--778},
  year={2016}
}

@misc{hu2021loralowrankadaptationlarge,
      title={LoRA: Low-Rank Adaptation of Large Language Models}, 
      author={Edward J. Hu and Yelong Shen and Phillip Wallis and Zeyuan Allen-Zhu and Yuanzhi Li and Shean Wang and Lu Wang and Weizhu Chen},
      year={2021},
      eprint={2106.09685},
      archivePrefix={arXiv},
      primaryClass={cs.CL},
      url={https://arxiv.org/abs/2106.09685}, 
}

@inproceedings{radford2021learning,
  title={Learning transferable visual models from natural language supervision},
  author={Radford, Alec and Kim, Jong Wook and Hallacy, Chris and Ramesh, Aditya and Goh, Gabriel and Agarwal, Sandhini and Sastry, Girish and Askell, Amanda and Mishkin, Pamela and Clark, Jack and others},
  booktitle={International conference on machine learning},
  pages={8748--8763},
  year={2021},
  organization={PMLR}
}

@inproceedings{zhai2023sigmoid,
  title={Sigmoid loss for language image pre-training},
  author={Zhai, Xiaohua and Mustafa, Basil and Kolesnikov, Alexander and Beyer, Lucas},
  booktitle={Proceedings of the IEEE/CVF International Conference on Computer Vision},
  pages={11975--11986},
  year={2023}
}

@inproceedings{papineni2002bleu,
  title={Bleu: a method for automatic evaluation of machine translation},
  author={Papineni, Kishore and Roukos, Salim and Ward, Todd and Zhu, Wei-Jing},
  booktitle={Proceedings of the 40th annual meeting of the Association for Computational Linguistics},
  pages={311--318},
  year={2002}
}

@inproceedings{lin2004rouge,
  title={Rouge: A package for automatic evaluation of summaries},
  author={Lin, Chin-Yew},
  booktitle={Text summarization branches out},
  pages={74--81},
  year={2004}
}

@inproceedings{banerjee2005meteor,
  title={METEOR: An automatic metric for MT evaluation with improved correlation with human judgments},
  author={Banerjee, Satanjeev and Lavie, Alon},
  booktitle={Proceedings of the acl workshop on intrinsic and extrinsic evaluation measures for machine translation and/or summarization},
  pages={65--72},
  year={2005}
}

@article{abdin2024phi,
  title={Phi-3 technical report: A highly capable language model locally on your phone},
  author={Abdin, Marah and Aneja, Jyoti and Awadalla, Hany and Awadallah, Ahmed and Awan, Ammar Ahmad and Bach, Nguyen and Bahree, Amit and Bakhtiari, Arash and Bao, Jianmin and Behl, Harkirat and others},
  journal={arXiv preprint arXiv:2404.14219},
  year={2024}
}

@article{yu2024rethinking,
  title={Rethinking vision transformer and masked autoencoder in multimodal face anti-spoofing},
  author={Yu, Zitong and Cai, Rizhao and Cui, Yawen and Liu, Xin and Hu, Yongjian and Kot, Alex C},
  journal={International Journal of Computer Vision},
  pages={1--22},
  year={2024},
  publisher={Springer}
}

@article{pietikainen2010local,
  title={Local binary patterns},
  author={Pietik{\"a}inen, Matti},
  journal={Scholarpedia},
  volume={5},
  number={3},
  pages={9775},
  year={2010}
}

@inproceedings{dalal2005histograms,
  title={Histograms of oriented gradients for human detection},
  author={Dalal, Navneet and Triggs, Bill},
  booktitle={2005 IEEE computer society conference on computer vision and pattern recognition (CVPR'05)},
  volume={1},
  pages={886--893},
  year={2005},
  organization={Ieee}
}

@article{yu2022deep,
  title={Deep learning for face anti-spoofing: A survey},
  author={Yu, Zitong and Qin, Yunxiao and Li, Xiaobai and Zhao, Chenxu and Lei, Zhen and Zhao, Guoying},
  journal={IEEE transactions on pattern analysis and machine intelligence},
  volume={45},
  number={5},
  pages={5609--5631},
  year={2022},
  publisher={IEEE}
}

@InProceedings{clip,
  title = 	 {Learning Transferable Visual Models From Natural Language Supervision},
  author =       {Radford, Alec and Kim, Jong Wook and Hallacy, Chris and Ramesh, Aditya and Goh, Gabriel and Agarwal, Sandhini and Sastry, Girish and Askell, Amanda and Mishkin, Pamela and Clark, Jack and Krueger, Gretchen and Sutskever, Ilya},
  booktitle = 	 ICML,
  pages = 	 {8748--8763},
  year = 	 {2021},
  volume = 	 {139},

}

@inproceedings{yu2019remote,
  title={Remote heart rate measurement from highly compressed facial videos: an end-to-end deep learning solution with video enhancement},
  author={Yu, Zitong and Peng, Wei and Li, Xiaobai and Hong, Xiaopeng and Zhao, Guoying},
  booktitle={Proceedings of the IEEE/CVF international conference on computer vision},
  pages={151--160},
  year={2019}
}

@article{qin2021meta,
  title={Meta-teacher for face anti-spoofing},
  author={Qin, Yunxiao and Yu, Zitong and Yan, Longbin and Wang, Zezheng and Zhao, Chenxu and Lei, Zhen},
  journal={IEEE transactions on pattern analysis and machine intelligence},
  volume={44},
  number={10},
  pages={6311--6326},
  year={2021},
  publisher={IEEE}
}

@inproceedings{icassp22multiple,
  author       = {Ying Bian and
                  Peng Zhang and
                  Jingjing Wang and
                  Chunmao Wang and
                  Shiliang Pu},
  title        = {Learning Multiple Explainable and Generalizable Cues for Face Anti-Spoofing},
  booktitle    = {{ICASSP}},
  pages        = {2310--2314},
  publisher    = {{IEEE}},
  year         = {2022}
}

@inproceedings{cvpr18aux,
  author       = {Yaojie Liu and
                  Amin Jourabloo and
                  Xiaoming Liu},
  title        = {Learning Deep Models for Face Anti-Spoofing: Binary or Auxiliary Supervision},
  booktitle    = {{CVPR}},
  pages        = {389--398},
  publisher    = {Computer Vision Foundation / {IEEE} Computer Society},
  year         = {2018}
}

@article{wang2025pnss,
  title={PNSS: Unknown Face Presentation Attack Detection with Pseudo Negative Sample Synthesis.},
  author={Wang, Hongyang and Shi, Yichen and Feng, Jun and Yu, Zitong and Tao, Zhuofu},
  journal={Computers, Materials \& Continua},
  volume={83},
  number={2},
  year={2025}
}

@article{lin2025reliable,
  title={Reliable and Balanced Transfer Learning for Generalized Multimodal Face Anti-Spoofing},
  author={Lin, Xun and Liu, Ajian and Yu, Zitong and Cai, Rizhao and Wang, Shuai and Yu, Yi and Wan, Jun and Lei, Zhen and Cao, Xiaochun and Kot, Alex},
  journal={IEEE Transactions on Pattern Analysis and Machine Intelligence},
  year={2025},
  publisher={IEEE}
}

@article{cai2025rehearsal,
  title={Rehearsal-free and efficient continual learning for cross-domain face anti-spoofing},
  author={Cai, Rizhao and Cui, Yawen and Yu, Zitong and Lin, Xun and Chen, Changsheng and Kot, Alex},
  journal={IEEE Transactions on Pattern Analysis and Machine Intelligence},
  year={2025},
  publisher={IEEE}
}

@inproceedings{flip,
  author       = {Koushik Srivatsan and
                  Muzammal Naseer and
                  Karthik Nandakumar},
  title        = {{FLIP:} Cross-domain Face Anti-spoofing with Language Guidance},
  booktitle    = {{ICCV}},
  pages        = {19628--19639},
  year         = {2023}
}

@inproceedings{eccv25tffas,
  author       = {Xudong Wang and
                  Ke{-}Yue Zhang and
                  Taiping Yao and
                  Qianyu Zhou and
                  Shouhong Ding and
                  Pingyang Dai and
                  Rongrong Ji},
  title        = {{TF-FAS:} Twofold-Element Fine-Grained Semantic Guidance for Generalizable
                  Face Anti-spoofing},
  booktitle    = {ECCV},
  volume       = {15065},
  pages        = {148--168},
  year         = {2024}
}

@inproceedings{cvpr24cfpl,
  author       = {Ajian Liu and
                  Shuai Xue and
                  Jianwen Gan and
                  Jun Wan and
                  Yanyan Liang and
                  Jiankang Deng and
                  Sergio Escalera and
                  Zhen Lei},
  title        = {{CFPL-FAS:} Class Free Prompt Learning for Generalizable Face Anti-Spoofing},
  booktitle    = {{CVPR}},
  pages        = {222--232},
  year         = {2024}
}

@article{tpami23survey,
  author       = {Zitong Yu and
                  Yunxiao Qin and
                  Xiaobai Li and
                  Chenxu Zhao and
                  Zhen Lei and
                  Guoying Zhao},
  title        = {Deep Learning for Face Anti-Spoofing: {A} Survey},
  journal      = {{IEEE} Trans. Pattern Anal. Mach. Intell.},
  volume       = {45},
  number       = {5},
  pages        = {5609--5631},
  year         = {2023}
}

@article{ijcv23adversarial,
  author       = {Fangling Jiang and
                  Qi Li and
                  Pengcheng Liu and
                  Xiang{-}Dong Zhou and
                  Zhenan Sun},
  title        = {Adversarial Learning Domain-Invariant Conditional Features for Robust
                  Face Anti-spoofing},
  journal      = {Int. J. Comput. Vis.},
  volume       = {131},
  number       = {7},
  pages        = {1680--1703},
  year         = {2023}
}

@inproceedings{cvpr24oneclass,
  author       = {Pei{-}Kai Huang and
                  Cheng{-}Hsuan Chiang and
                  Tzu{-}Hsien Chen and
                  Jun{-}Xiong Chong and
                  Tyng{-}Luh Liu and
                  Chiou{-}Ting Hsu},
  title        = {One-Class Face Anti-Spoofing via Spoof Cue Map-Guided Feature Learning},
  booktitle    = {{CVPR}},
  pages        = {277--286},
  year         = {2024}
}

@inproceedings{cvpr24gradient,
  author       = {Binh Minh Le and
                  Simon S. Woo},
  title        = {Gradient Alignment for Cross-Domain Face Anti-Spoofing},
  booktitle    = {CVPR},
  pages        = {188--199},
  year         = {2024}
}

@inproceedings{cvpr22ssan,
  author       = {Zhuo Wang and
                  Zezheng Wang and
                  Zitong Yu and
                  Weihong Deng and
                  Jiahong Li and
                  Tingting Gao and
                  Zhongyuan Wang},
  title        = {Domain Generalization via Shuffled Style Assembly for Face Anti-Spoofing},
  booktitle    = {{CVPR}},
  pages        = {4113--4123},
  year         = {2022}
}

@article{ijcv24augmentation,
  author       = {Rizhao Cai and
                  Cecelia Soh and
                  Zitong Yu and
                  Haoliang Li and
                  Wenhan Yang and
                  Alex C. Kot},
  title        = {Towards Data-Centric Face Anti-Spoofing: Improving Cross-domain Generalization
                  via Physics-based Data Synthesis},
  journal      = IJCV,
  year         = {2024}
}

@inproceedings{aaai22generation,
  author       = {Shice Liu and
                  Shitao Lu and
                  Hongyi Xu and
                  Jing Yang and
                  Shouhong Ding and
                  Lizhuang Ma},
  title        = {Feature Generation and Hypothesis Verification for Reliable Face Anti-spoofing},
  booktitle    = {{AAAI}},
  pages        = {1782--1791},
  year         = {2022}
}

@article{gpt4,
  title={Gpt-4 technical report},
  author={Achiam, Josh and others},
  journal={arXiv preprint arXiv:2303.08774},
  year={2023}
}

@inproceedings{huang2024aesexpert,
  title={Aesexpert: Towards multi-modality foundation model for image aesthetics perception},
  author={Huang, Yipo and Sheng, Xiangfei and Yang, Zhichao and Yuan, Quan and Duan, Zhichao and Chen, Pengfei and Li, Leida and Lin, Weisi and Shi, Guangming},
  booktitle={Proceedings of the 32nd ACM International Conference on Multimedia},
  pages={5911--5920},
  year={2024}
}

@misc{li2023seedbench2benchmarkingmultimodallarge,
      title={SEED-Bench-2: Benchmarking Multimodal Large Language Models}, 
      author={Bohao Li and Yuying Ge and Yixiao Ge and Guangzhi Wang and Rui Wang and Ruimao Zhang and Ying Shan},
      year={2023},
      eprint={2311.17092},
      archivePrefix={arXiv},
      primaryClass={cs.CV},
      url={https://arxiv.org/abs/2311.17092}, 
}

@misc{huang2024aesbenchexpertbenchmarkmultimodal,
      title={AesBench: An Expert Benchmark for Multimodal Large Language Models on Image Aesthetics Perception}, 
      author={Yipo Huang and Quan Yuan and Xiangfei Sheng and Zhichao Yang and Haoning Wu and Pengfei Chen and Yuzhe Yang and Leida Li and Weisi Lin},
      year={2024},
      eprint={2401.08276},
      archivePrefix={arXiv},
      primaryClass={cs.CV},
      url={https://arxiv.org/abs/2401.08276}, 
}

@misc{song2024moresimpleeffectivetoken,
      title={Less is More: A Simple yet Effective Token Reduction Method for Efficient Multi-modal LLMs},
      author={Dingjie Song and Wenjun Wang and Shunian Chen and Xidong Wang and Michael Guan and Benyou Wang},
      year={2024},
      eprint={2409.10994},
      archivePrefix={arXiv},
      primaryClass={cs.CL},
      url={https://arxiv.org/abs/2409.10994},
}

@article{dubey2024llama,
  title={The llama 3 herd of models},
  author={Dubey, Abhimanyu and Jauhri, Abhinav and Pandey, Abhinav and Kadian, Abhishek and Al-Dahle, Ahmad and Letman, Aiesha and Mathur, Akhil and Schelten, Alan and Yang, Amy and Fan, Angela and others},
  journal={arXiv preprint arXiv:2407.21783},
  year={2024}
}

@article{zhou2022learning,
  title={Learning to prompt for vision-language models},
  author={Zhou, Kaiyang and Yang, Jingkang and Loy, Chen Change and Liu, Ziwei},
  journal={International Journal of Computer Vision},
  volume={130},
  number={9},
  pages={2337--2348},
  year={2022},
  publisher={Springer}
}

@inproceedings{zhou2023instance,
  title={Instance-aware domain generalization for face anti-spoofing},
  author={Zhou, Qianyu and Zhang, Ke-Yue and Yao, Taiping and Lu, Xuequan and Yi, Ran and Ding, Shouhong and Ma, Lizhuang},
  booktitle={Proceedings of the IEEE/CVF conference on computer vision and pattern recognition},
  pages={20453--20463},
  year={2023}
}

@article{zhang2025interpretable,
  title={Interpretable Face Anti-Spoofing: Enhancing Generalization with Multimodal Large Language Models},
  author={Zhang, Guosheng and Wang, Keyao and Yue, Haixiao and Liu, Ajian and Zhang, Gang and Yao, Kun and Ding, Errui and Wang, Jingdong},
  journal={arXiv preprint arXiv:2501.01720},
  year={2025}
}

@article{peng2023kosmos,
  title={Kosmos-2: Grounding multimodal large language models to the world},
  author={Peng, Zhiliang and Wang, Wenhui and Dong, Li and Hao, Yaru and Huang, Shaohan and Ma, Shuming and Wei, Furu},
  journal={arXiv preprint arXiv:2306.14824},
  year={2023}
}

\clearpage
\appendix
\setcounter{page}{1}

\title{\modelname}
\section{Details about FaccShield Dataset}
\label{sec:dataset}

\subsection{Data Source}


The FaceShield dataset, comprising FaceShield-pre10K for pretraining and FaceShield-sft45K for supervised fine-tuning (SFT), is constructed using images sourced from three widely recognized datasets: WMCA~\cite{8714076}, SiW-Mv2~\cite{xiaoguo2022MDFAS}, and PADISI~\cite{Rostami_2021_ICCV}. These foundational datasets provide a diverse basis for generating question-answer (QA) pairs that address a variety of tasks in anti-spoofing research.

In the pretraining phase, the FaceShield-pre10K dataset focuses on generating QA pairs that describe the visual content of the images. A total of 9,297 QA pairs were created to establish a foundational understanding of the visual data. These pairs enable models to grasp core visual concepts and perform basic reasoning related to spoofing scenarios, forming the groundwork for subsequent learning.

In contrast, the FaceShield-sft45K dataset, used in the SFT phase, incorporates task-specific annotations to address more complex objectives. This dataset contains 45,662 QA pairs, which are categorized into several distinct tasks: coarse-grained classification, fine-grained classification, reasoning, and attack localization. These QA pairs are designed to meet practical anti-spoofing requirements, enabling models to handle diverse challenges across various anti-spoofing tasks.

\begin{figure*}[t]
    \centering
    \vspace{-1.3em}
    \begin{subfigure}{0.48\linewidth} 
        \centering
        \includegraphics[width=\linewidth]{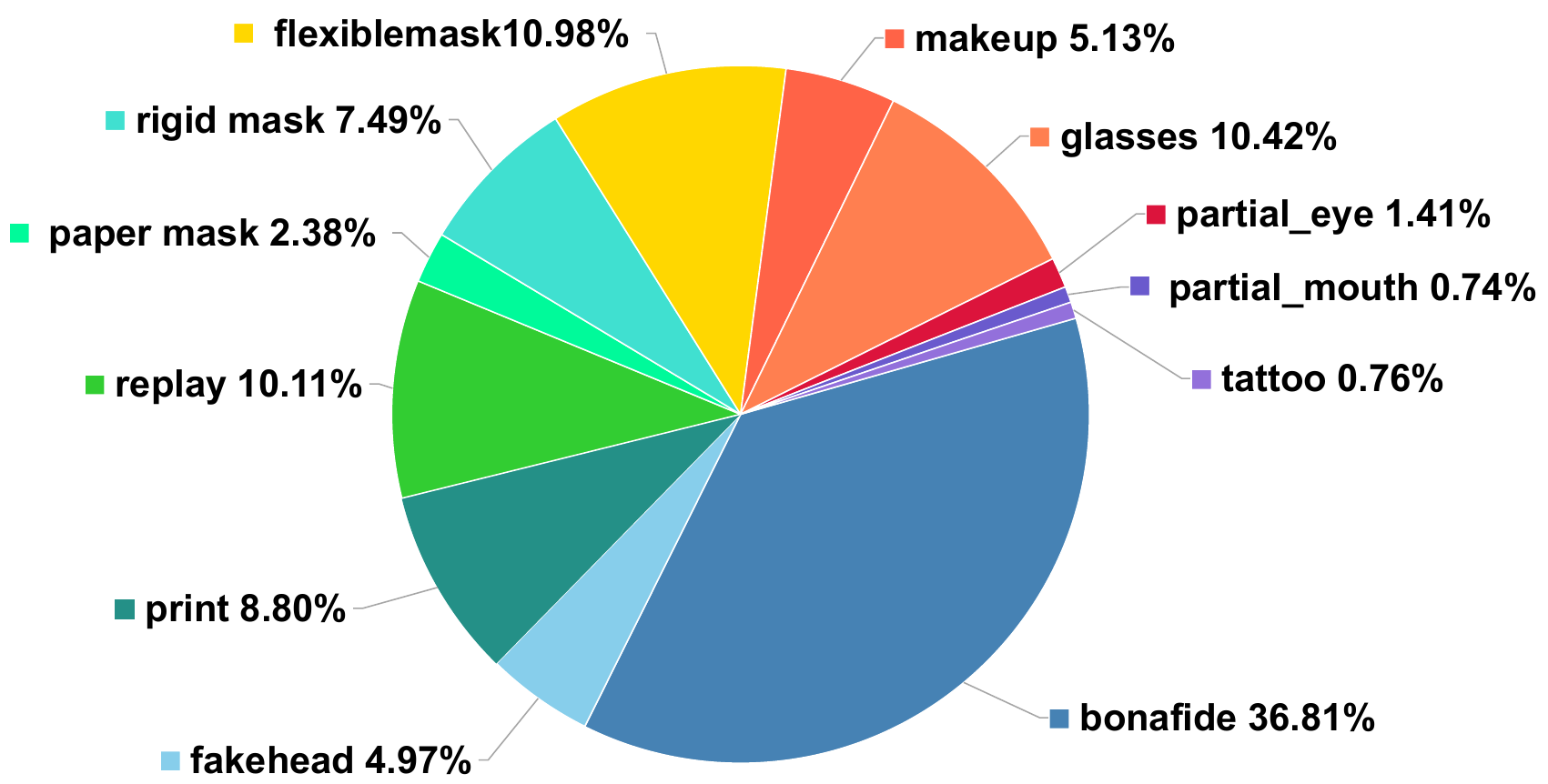}
        \vspace{-0.8em}
        \label{fig:pretrain_pie}
    \end{subfigure}
    \hfill
    \begin{subfigure}{0.48\linewidth} 
        \centering
        \includegraphics[width=\linewidth]{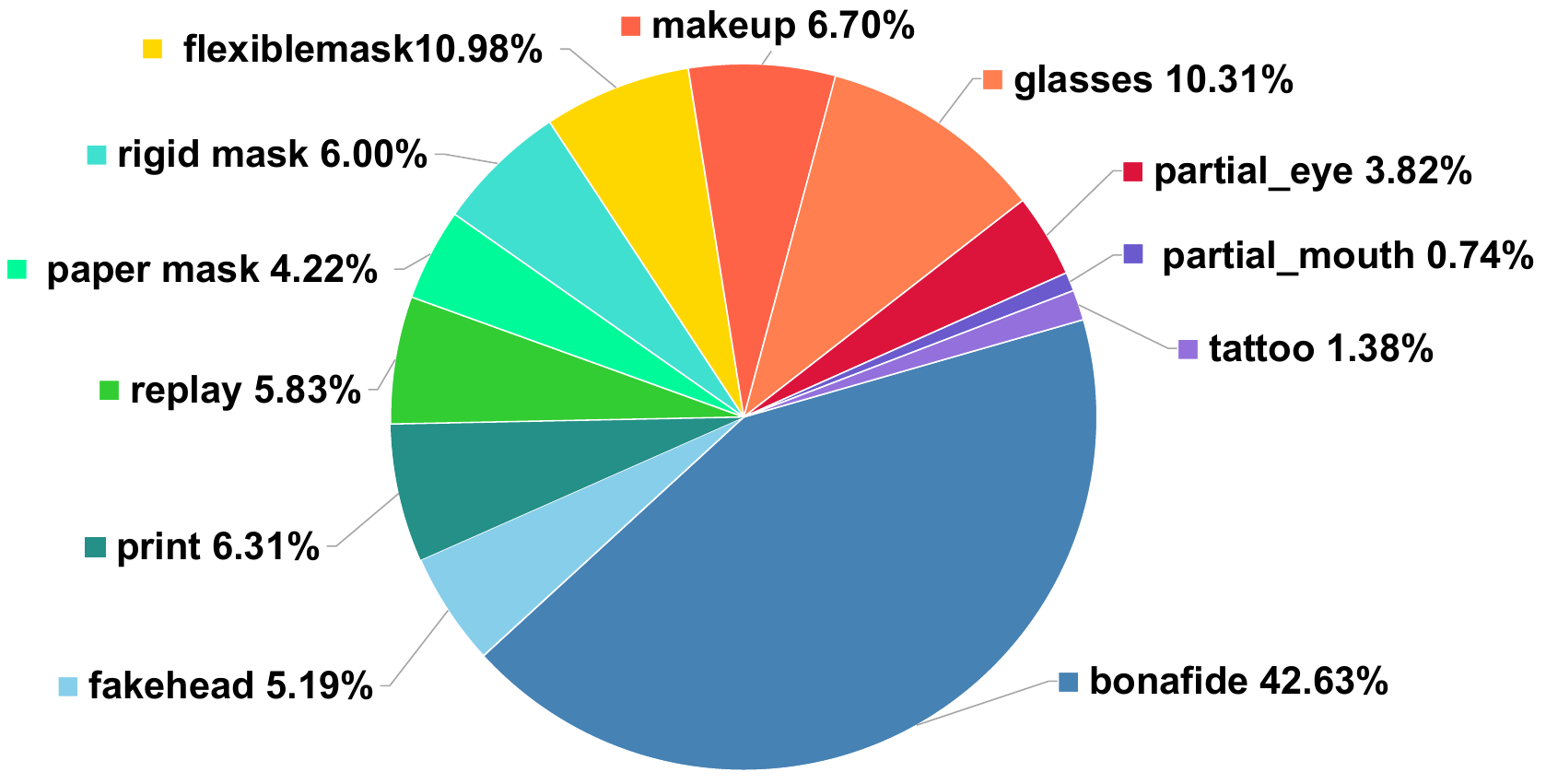}
        \vspace{-0.8em}
        \label{fig:sft_pie}
    \end{subfigure}
    \caption{Comparison of category distributions in \datasetpre\ and \datasetsft\ datasets.}
    \label{fig:combined_piecharts}
\end{figure*}

The FaceShield-pre10K dataset emphasizes descriptive comprehension and fundamental reasoning about image content, while the FaceShield-sft45K dataset advances to specialized, task-driven annotations. The distribution of QA pairs is detailed in Table \ref{tab:DatasetStatistics} and the category distributions in Fig.~\ref{fig:combined_piecharts}. 

Table.~\ref{tab:Comparison of datasets} compares face anti-spoofing datasets. Unlike traditional datasets with only binary labels for print and replay attacks, FaceShield provides richer annotations (including attack types and bounding boxes) and supports multimodal tasks with large-scale image-caption and QA pairs.

\begin{table}[H]
\centering
\caption{Dataset Statistics(QA pairs)}
\label{tab:DatasetStatistics}
\resizebox{0.95\columnwidth}{!}{%
\begin{tabular}{lccc}
\hline
\textbf{Datasets}       & \textbf{Source} & \textbf{Data Size} & \textbf{Total} \\ \hline
                        & WMCA~\cite{8714076}           & 3875               &                \\
\datasetpre                 & SiW-Mv2~\cite{xiaoguo2022MDFAS}             & 3782               &    9297            \\
                        & PADISI~\cite{Rostami_2021_ICCV}          & 1640               &            \\ \hline
                        & WMCA~\cite{8714076}            & 16776              &                \\
\datasetsft                 & SiW-Mv2~\cite{xiaoguo2022MDFAS}             & 18096              &      45662          \\
                        & PADISI~\cite{Rostami_2021_ICCV}          & 10790              &           \\ \hline
\end{tabular}%
}
\end{table}

\vspace{-2.5mm}
\begin{table}[h]
\centering
\caption{Comparison of existing datasets}
\vspace{-2.5mm}
\resizebox{\columnwidth}{!}{
\begin{tabular}{l l l}
\toprule
\textbf{Dataset} & \textbf{Attack Types} & \textbf{Annotations} \\
\midrule
SiW        & Print, Replay(2)              & Binary class \\
Oulu-NPU        & Print, Replay(2)              & Binary class \\
CASIA-MFSD      & Print, Replay(2)              & Binary class \\
Replay-Attack   & Print, Replay(2)             & Binary class \\
MSU             & Print, Replay(2)              & Binary class \\
ROSE             & Print, Replay, PaperMask(3)              & Binary class \\
FaceShield      & Unified-attack (11 types)  & QA and bounding boxes \\
\bottomrule
\end{tabular}
}
\label{tab:Comparison of datasets}
\end{table}
\vspace{-2.5mm}

\subsection{Data Format}
\begin{figure}[H]
    \centering
    \vspace{-0.8em}
    \includegraphics[width=0.85\linewidth]{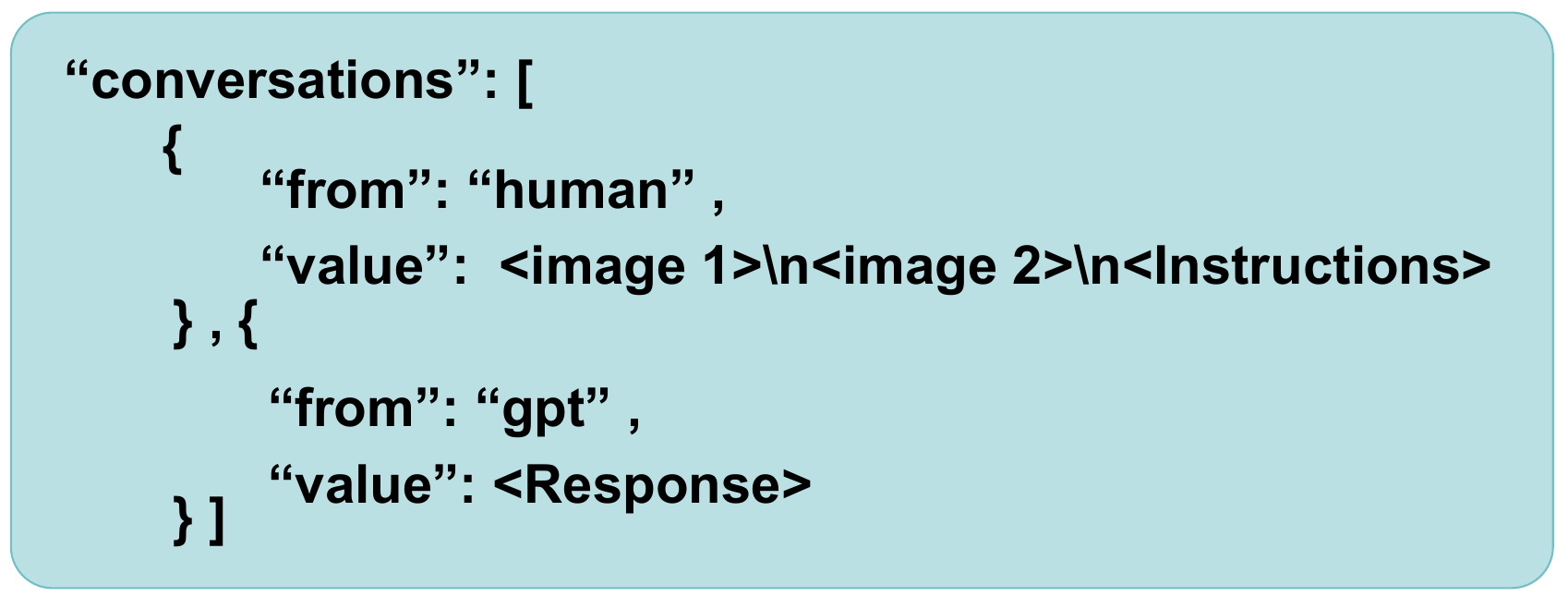}
    \vspace{-0.8em}
    \caption{Data format}
    \label{fig:dataformat}
    \vspace{-1.4em}
\end{figure}

We constructed the dataset based on the LLAVA framework, as shown in Fig.~\ref{fig:dataformat}. 

\subsection{Details about \datasetpre}
\label{sec:datasetpre}

\subsubsection{Data Generation}
\begin{figure}[H]
    \centering
    \includegraphics[width=1\linewidth]{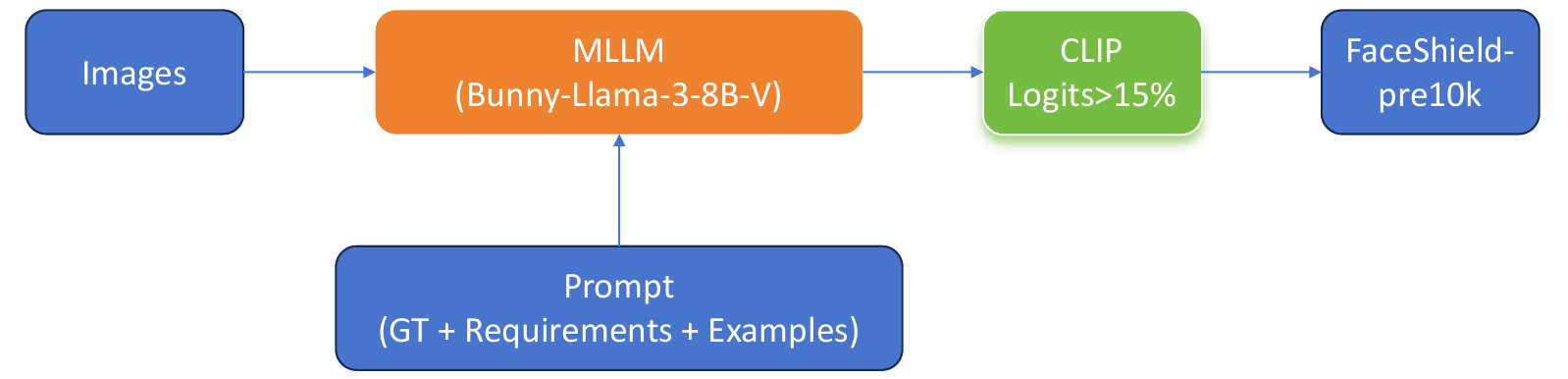}
     \vspace{-0.8em}
    \caption{\datasetpre construction pipeline}
    \label{fig:predata_pipeline}
\end{figure}

As shown in Fig.~\ref{fig:predata_pipeline}, we employed Bunny-Llama-3-8B-V (MLLM)~\cite{he2024bunny} to generate high-quality question-answer (QA) pairs aimed at describing the visual content of images. The primary objective of this dataset is to equip the MLLM with a foundational understanding of visual attributes relevant to the Face Anti-Spoofing (FAS) task, enabling it to develop basic perceptual capabilities to distinguish between real and spoofed faces.

The QA generation process is guided by a systematically designed prompt framework comprising three key components. First, each image is paired with its corresponding Ground Truth (GT), which denotes the spoofing category (e.g., glasses, mask, makeup). The GT serves as contextual information to ensure that the generated QA pairs are semantically aligned with the spoofing type depicted in the image. Second, the construction requirements are explicitly defined to standardize the format and content of the QA pairs. Each question is concise, typically a single sentence, and designed to elicit descriptions focused on either the environment (e.g., background, lighting) or facial attributes (e.g., gender, facial organ, expression)~\cite{eccv25tffas}. The answer, by contrast, is required to be detailed and comprehensive, integrating observations about both the environment and facial features. This approach ensures that the QA pairs are both informative and task-relevant. Lastly, the prompts include illustrative examples to provide clear guidance on the desired output structure and level of detail. These examples demonstrate the expected question format, the richness required in the answers, and the integration of diverse attributes such as lighting conditions and facial expressions. This structured approach ensures that the generated QA pairs are of high quality, semantically relevant, and tailored to the needs of the FAS domain. Consequently, the FaceShield-pre10K dataset provides a robust foundation for the MLLM to develop perceptual capabilities, facilitating its adaptation to more complex and specialized tasks in the subsequent fine-tuning phase.

\subsubsection{Data Filtering}

To ensure the quality of the generated QA pairs, we utilized the CLIP~\cite{clip} model to evaluate the semantic alignment between the images and their corresponding textual descriptions. The filtering process was based on similarity scores computed between each image and its associated text. Samples with a similarity score below 15\% were considered low-quality and excluded from the dataset, while the rest were retained.

For text data exceeding the 77-token limit of CLIP~\cite{clip}, the text was segmented into multiple sentences, and similarity scores were computed for each sentence. The final similarity score was calculated as the average of the sentence-level scores. For text samples within the token limit, the similarity score was directly computed using the entire text. This approach ensured accurate evaluation for longer text samples while maintaining consistency for shorter ones.

After applying this filtering process, the dataset was reduced from 12091 to 9297 QA pairs. This step ensured that the dataset retained only high-quality, semantically consistent data, forming a reliable basis for pretraining tasks in the FAS domain.

\subsection{Details about \datasetsft}
\label{sec:datasetsft}

\begin{figure}[H]
    \centering
    \vspace{-1.3em}
    \includegraphics[width=1\linewidth]{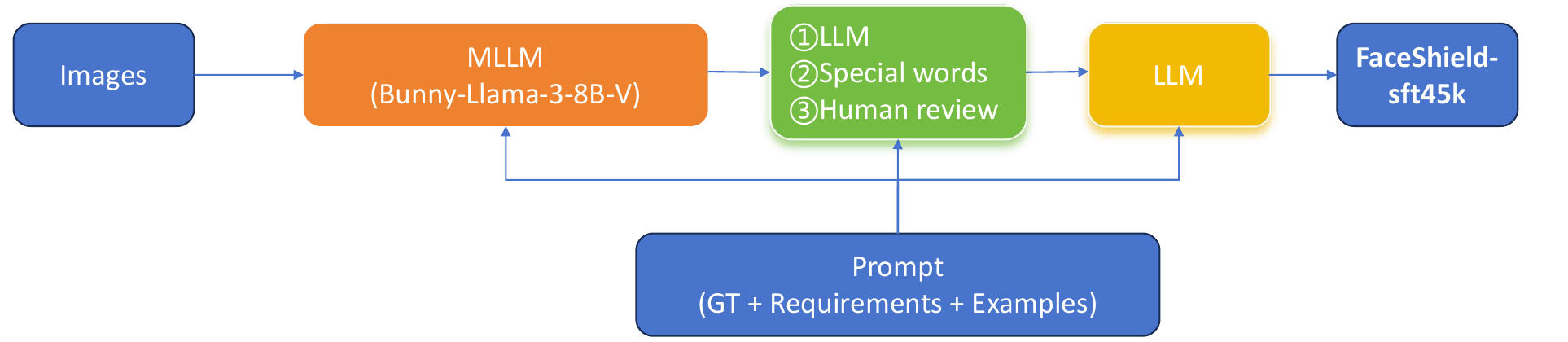}
     \vspace{-0.8em}
    \caption{\datasetsft\ construction pipeline}
    \label{fig:sftdaaset_pipeline}
\end{figure}

As shown in Fig.~\ref{fig:sftdaaset_pipeline}, we utilized Bunny-Llama-3-8B-V~\cite{he2024bunny} to generate high-quality QA pairs for four tasks: Coarse-grained Classification, Fine-grained Classification, Reasoning, and Attack Localization. The prompts for data generation follow a common structure comprising Ground Truth (GT), Requirements, and Examples. GT provides contextual information (e.g., real or spoof labels, spoof types, and bounding box coordinates), Requirements specify the desired QA format, and Examples guide generation with representative samples. While the structure is consistent, the content is tailored to the objectives of each task. Data sizes for each task are detailed in Table~\ref{table:faceshield_data}.

The construction process, outlined in Algorithm~\ref{alg:FaceShield-sft45K}, includes generating QA pairs using multimodal input, quality filtering, human review, and data augmentation through rephrasing. This pipeline ensures a diverse, high-quality dataset optimized for the specified tasks.

\begin{table}[H]
\caption{Task and Data Size(QA pairs) for FaceShield-sft45K Dataset}
\centering
\small 
\begin{tabular}{@{}ccc@{}}
\toprule
\textbf{Dataset}         & \textbf{Task}                      & \textbf{Data Size} \\ \midrule
\multirow{4}{*}{\centering FaceShield-sft45K} 
                         & Coarse-grained Classification      & 13076              \\ 
                         & Fine-grained Classification        & 12749              \\ 
                         & Reasoning                          & 11082              \\ 
                         & Attack Localization                & 8755               \\ \bottomrule
\end{tabular}
\label{table:faceshield_data}
\end{table}

\subsubsection{Multi-task Data generation}
\begin{figure}[t]
    \centering
    \includegraphics[width=1\linewidth]{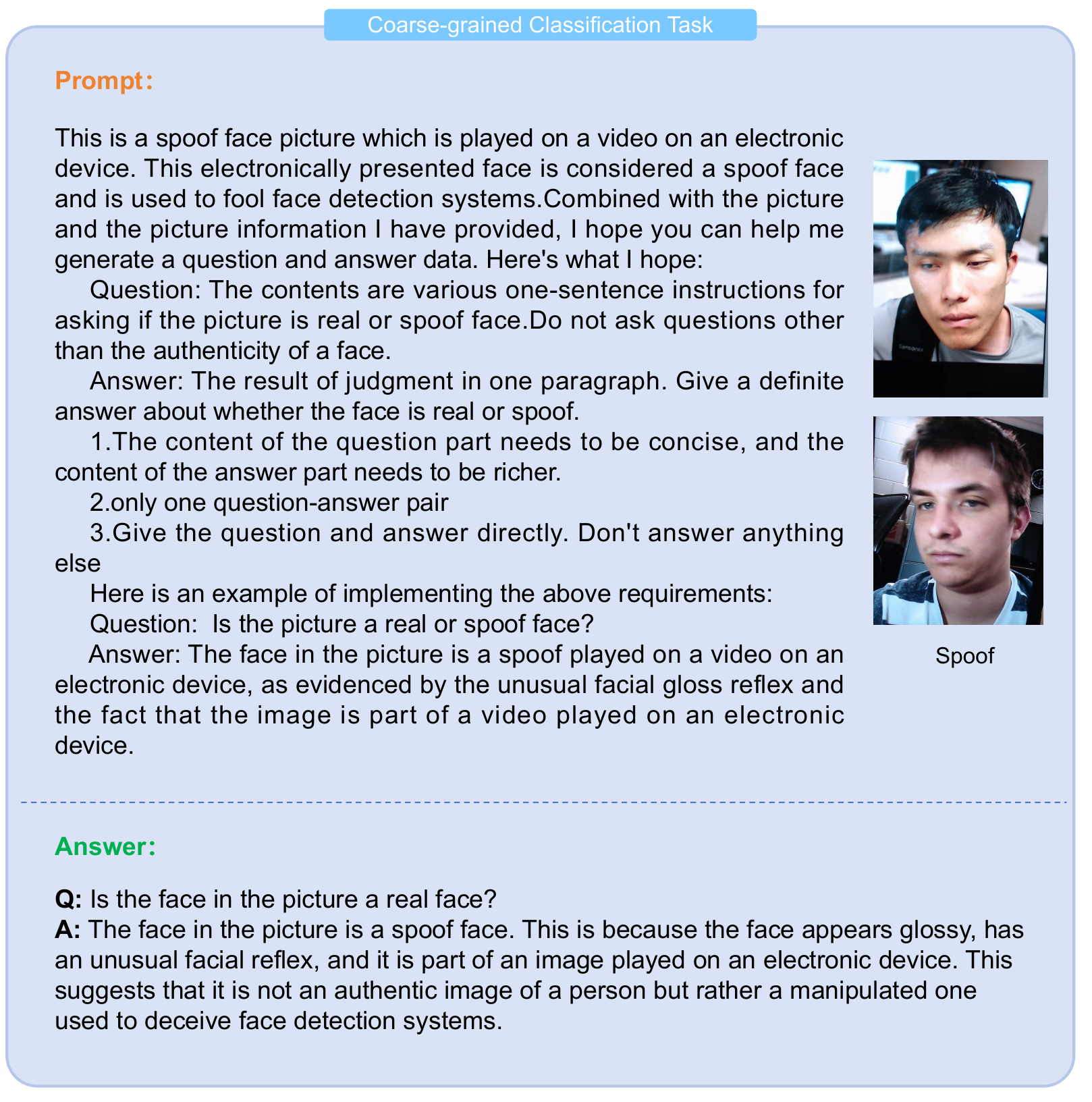}
     \vspace{-0.8em}
    \caption{Prompt for Coarse-grained Classification Task}
    \label{fig:Prompt_co_class}
\end{figure}

\paragraph{Coarse-grained Classification Task}
The coarse-grained classification task aims to determine whether a face in an image is real or spoofed. This task addresses the fundamental binary classification problem central to FAS. In this task, the prompts focus on eliciting clear and definitive judgments about the authenticity of the face, supported by reasoning tied to visual features. For example, like Fig.~\ref{fig:Prompt_co_class}, a typical prompt requires the MLLM to generate a question such as, ``Is the face in the picture real or spoof?'' and an answer like, ``The face in the picture is a spoof face. This is because the face appears glossy, has an unusual facial reflex, and therefore it is part of an image played on an electronic device.'' This task provides the model with foundational knowledge to identify face authenticity.

\begin{figure}[t]
    \centering
    \vspace{-1.3em}
    \includegraphics[width=1\linewidth]{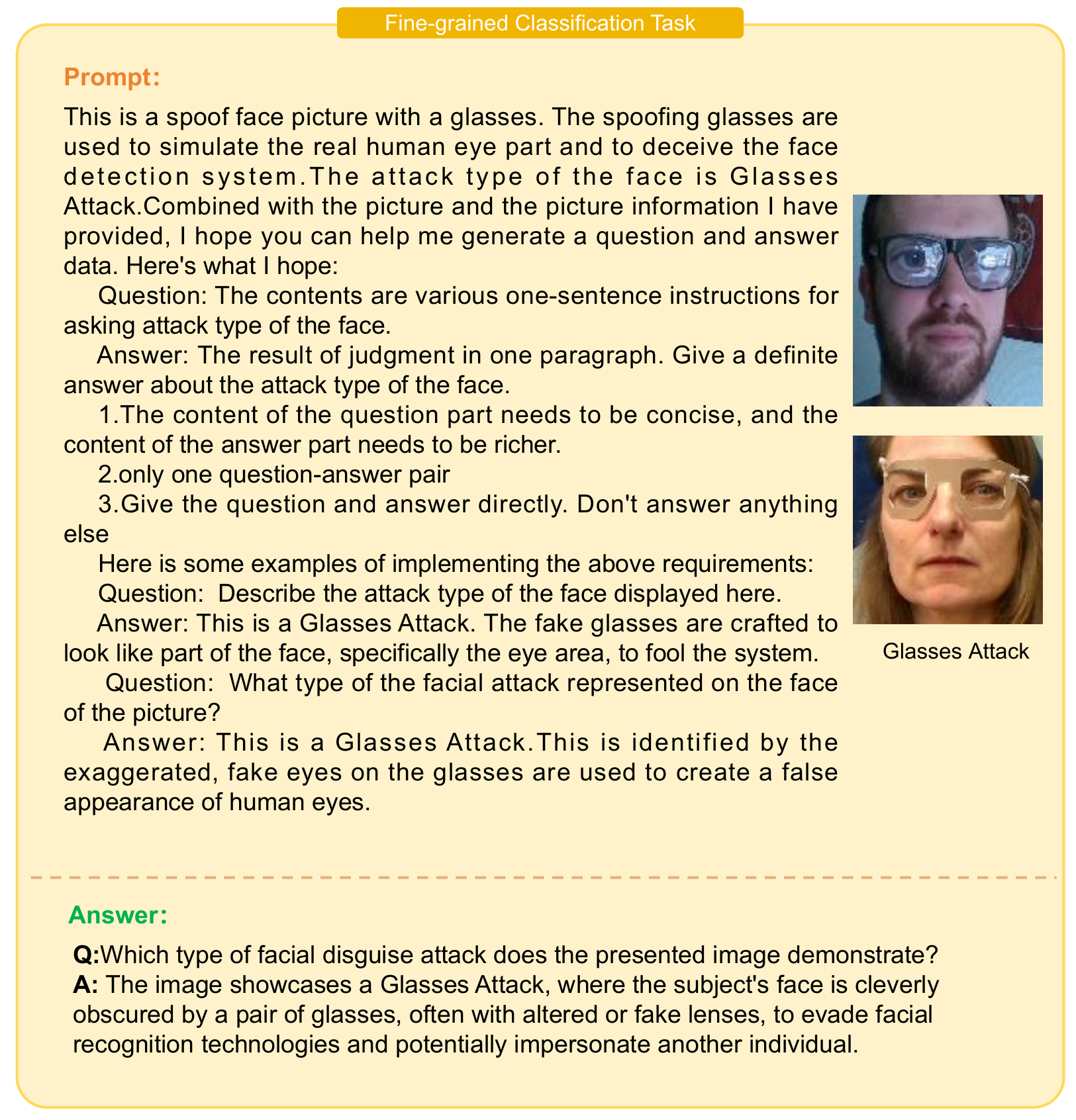}
     \vspace{-0.8em}
    \caption{Prompt for Fine-grained Classification Task}
    \label{fig:Prompt_fine_class}
\end{figure}

\paragraph{Fine-grained Classification Task}
The fine-grained classification task extends the binary classification by requiring the model to identify not only whether the face is real or spoofed but also the specific spoof type (e.g., glasses, fakehead, makeup attack). The prompts for this task are designed to ensure the generated QA pairs explicitly address these finer distinctions. For example, like Fig.~\ref{fig:Prompt_fine_class}, a question might ask, ``What attack type of facial spoof is shown in this image?'' with an answer such as, ``The image showcases a Glasses Attack, where the subject's face is cleverly obscured by a pair of glasses.'' This task enhances the model's ability to detect and classify diverse spoofing techniques.

\begin{figure}[t]
    \centering
    \vspace{-1.3em}
    \includegraphics[width=1\linewidth]{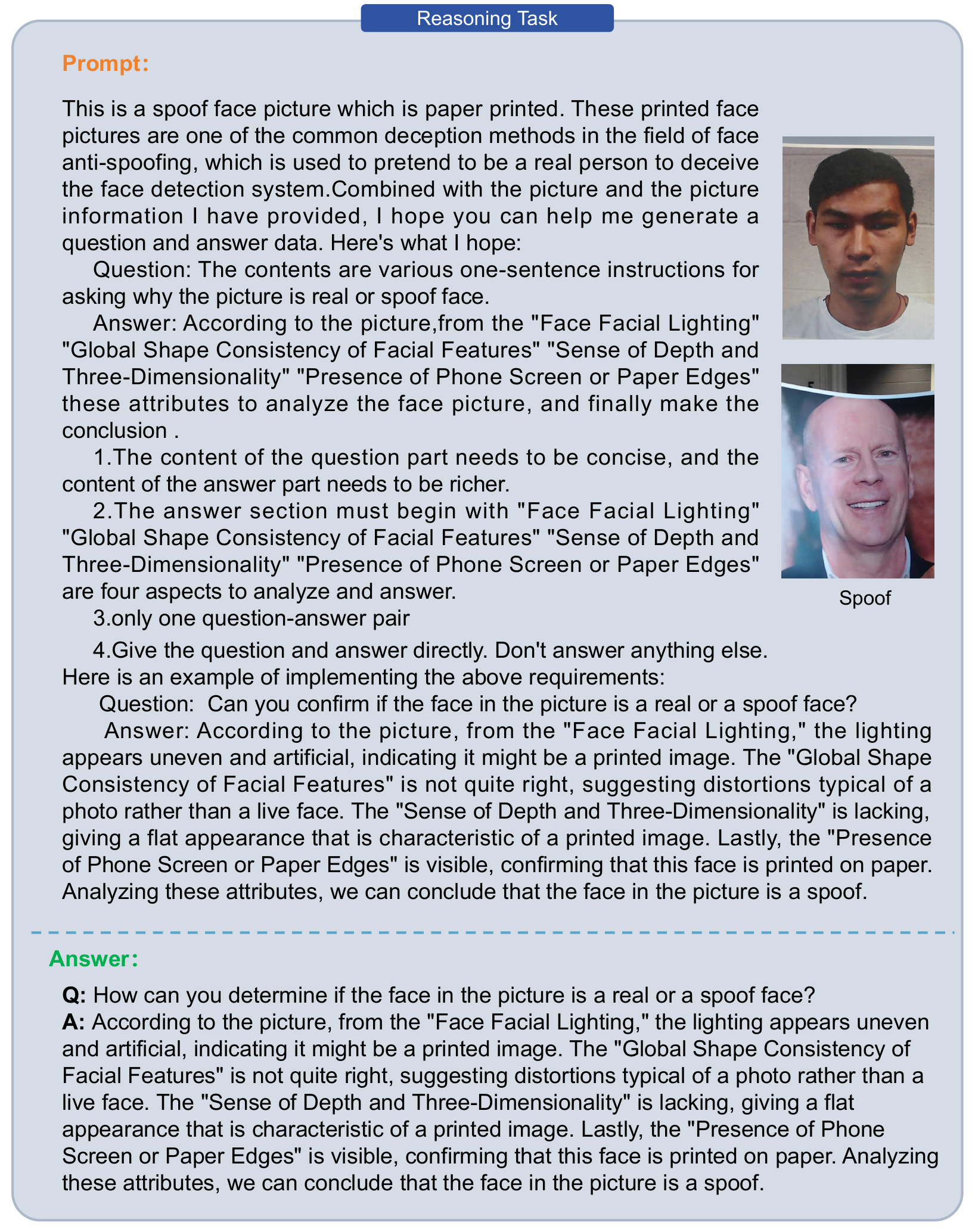}
     \vspace{-0.8em}
    \caption{Prompt for Reasoning Task}
    \label{fig:Prompt_reasoning}
\end{figure}

\paragraph{Reasoning Task}
The reasoning task is designed to enhance the model's ability to provide structured and interpretable explanations for its judgments, fostering a deeper understanding of FAS scenarios. For instance, like Fig.~\ref{fig:Prompt_reasoning}, the question might be phrased as: ``Why is the face real or spoofed?'' The answer requires the model to systematically evaluate the image based on four key attributes: Facial Lighting, Global Shape Consistency of Facial Features, Sense of Depth and Three-Dimensionality, and Presence of Phone Screen or Paper Edges. These attributes are selected because they represent core indicators in FAS tasks. Facial Lighting highlights irregularities caused by artificial materials or spoofing media under light. Global Shape Consistency assesses whether the facial features align naturally, a key distinction often disrupted in spoofing attempts. Sense of Depth and Three-Dimensionality captures the lack of realistic depth commonly found in spoofed faces. Presence of Phone Screen or Paper Edges identifies visible edges or surfaces indicative of physical spoofing mediums, such as photographs or displays. By methodically analyzing each of these attributes and integrating their observations, the model delivers a comprehensive judgment on the authenticity of the face.

\begin{figure}[t]
    \centering
    \vspace{-1.3em}
    \includegraphics[width=1\linewidth]{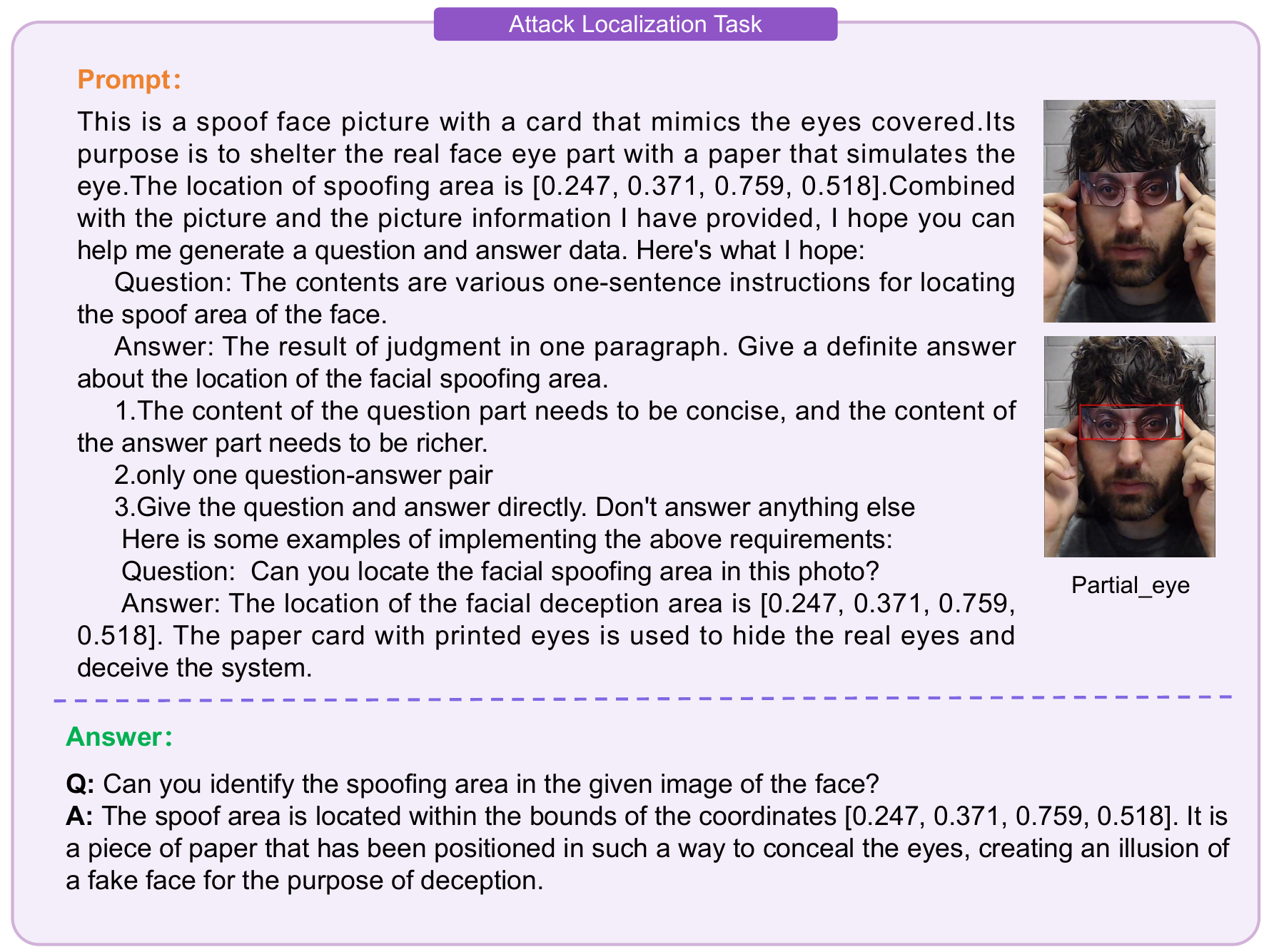}
     \vspace{-0.8em}
    \caption{Prompt for Attack Localization Task}
    \label{fig:Prompt_localization}
\end{figure}

\paragraph{Attack Localization Task}  
The attack localization task focuses on identifying the specific region of a spoofing attack within an image, which is critical for detailed analysis and interpretability in FAS scenarios. The prompts require the model to determine whether a face is real or spoofed and, if spoofed, to locate the spoofing region. For instance, like Fig.~\ref{fig:Prompt_localization}, the generated QA might include a question like, ``Can you locate the spoof area of the face in the image?'' The corresponding answer should provide the bounding box (bbox) coordinates of the spoofing region (e.g., eyes, mouth) and a concise description of the observed spoofing characteristics. This task ensures that the model not only identifies the presence of a spoof but also pinpoints its location, thereby improving interpretability in FAS decision-making.

\subsubsection{Data Filtering}
To ensure the quality and task relevance of the QA pairs in the FaceShield-sft45K dataset, a multi-stage filtering process was employed, incorporating LLM-based evaluation, special words filtering, and human review. This pipeline ensured that low-quality or irrelevant QA pairs were systematically identified and removed, resulting in a high-quality dataset suitable for supervised fine-tuning in Face Anti-Spoofing (FAS) tasks.

\paragraph{\textbf{LLM-based Filtering}}  
The first stage of filtering utilizes an LLM to evaluate the semantic consistency between each question and its corresponding answer. The LLM was tasked with determining whether the answer directly addressed the question and adhered to the specific requirements of the task. QA pairs were flagged for exclusion if:
\begin{enumerate}
    \item The question and answer exhibited semantic misalignment or inconsistency.
    \item The answer deviated from the task objective or provided irrelevant information.
\end{enumerate}
This automated evaluation allowed us to efficiently filter out QA pairs that lacked coherence or failed to meet the standards of task relevance.

\paragraph{\textbf{Special Words Filtering}}  
In parallel, a second filtering mechanism was implemented to identify QA pairs containing predefined special words or phrases indicative of low-quality responses. These phrases typically signaled either the model’s inability to provide a meaningful answer or responses irrelevant to the task. Examples of such phrases include:
\begin{itemize}
    \item \textit{“As an AI language model, I cannot...”}
    \item \textit{“It is difficult to determine...”}
    \item \textit{“It is not possible to...”}
\end{itemize}
QA pairs containing such phrases were automatically removed, as they failed to provide informative or useful annotations for the dataset.

\paragraph{\textbf{Human Review and Refinement}}  
Following the automated filtering stages, human reviewers conducted a final review of the retained QA pairs to ensure quality and consistency. This review process focused on:
\begin{enumerate}
    \item Addressing cases where automated filtering may have incorrectly flagged or missed low-quality QA pairs.
    \item Correcting minor errors, such as grammatical inconsistencies or subtle misalignments between questions and answers.
    \item Refining edge cases that required nuanced contextual understanding.
\end{enumerate}
The human review provided an additional layer of quality assurance, ensuring that the dataset met the required standards of accuracy, relevance, and clarity.

\subsubsection{Data Augmentation}

\begin{figure}[t]
    \centering
    \vspace{-1.3em}
    \includegraphics[width=1\linewidth]{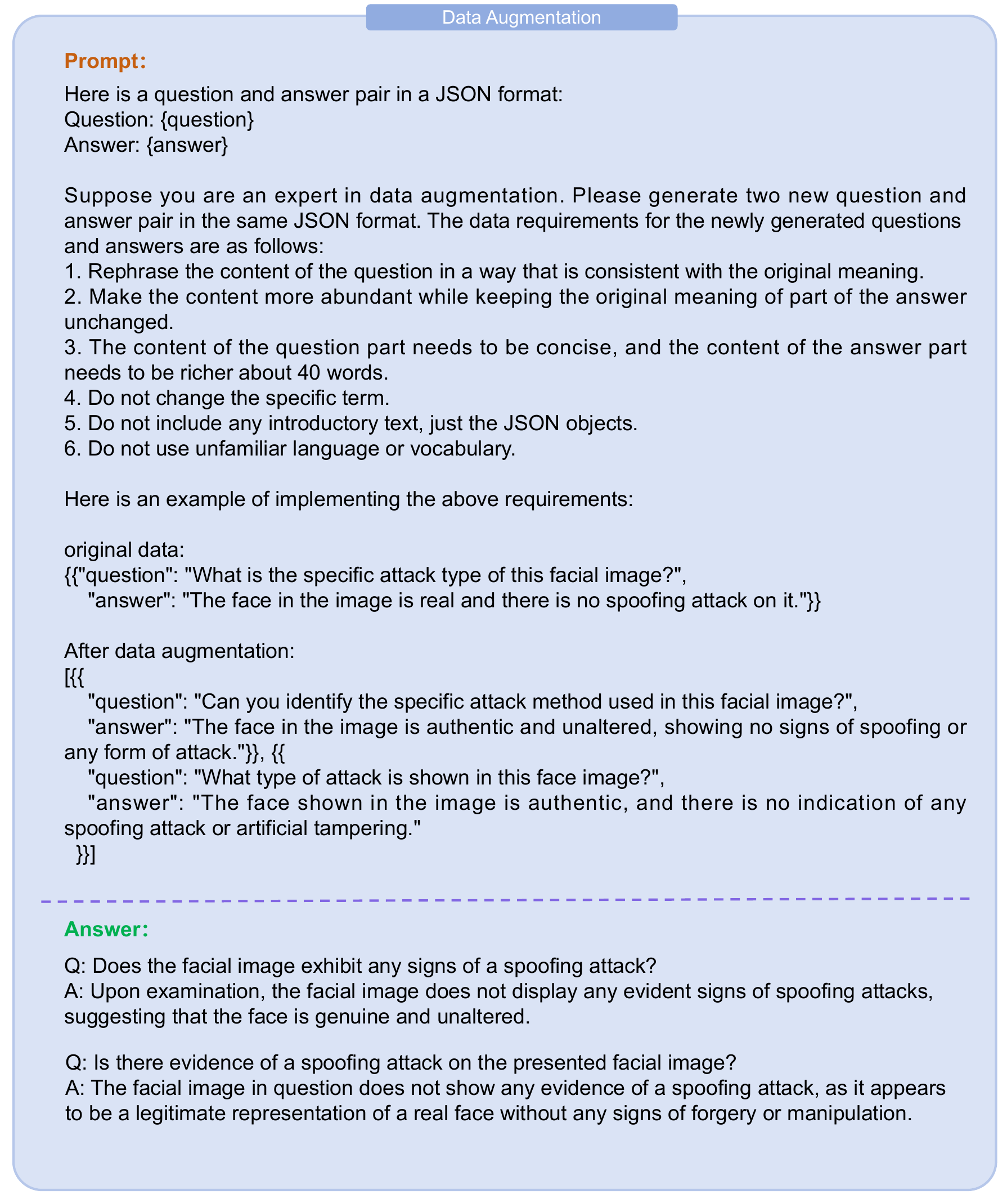}
     \vspace{-0.8em}
    \caption{Prompt for Data augmentation}
    \label{fig:Prompt_aug}
\end{figure}

To further increase the diversity and quantity of QA pairs in the FaceShield-sft45K dataset, a systematic data augmentation strategy that leverages the generative capabilities of the LLM was employed. As shown in Fig.~\ref{fig:Prompt_aug}, this method used the existing QA pairs as input and generated additional semantically equivalent or closely related QA pairs, enhancing linguistic variability while maintaining the original semantic intent.

The augmentation process was structured as follows: the LLM was provided with a QA pair in a structured format, including both the question and answer. The prompt instructed the LLM to act as an expert in data augmentation, generating multiple new QA pairs that conveyed the same meaning as the original but differed in phrasing, word choice, and structural elements. The generated outputs were designed to reflect diverse linguistic expressions while ensuring semantic alignment with the original QA pair.

By applying this augmentation process consistently across the dataset, the FaceShield-sft45K dataset was significantly enriched with variations that introduced natural linguistic diversity. This augmentation ensured that the dataset could support robust model training by exposing the models to a wider range of expressions and syntactic patterns, improving their ability to generalize across varied language inputs.

\subsection{Ablation Study}

This section investigates the contributions of key components in the training pipeline, specifically the inclusion of the FaceShield-pre10K dataset during pretraining and the data filtering and augmentation processes during fine-tuning. The ablation experiments provide quantitative insights into the significance of these components.

To evaluate the impact of pretraining on the FaceShield-pre10K dataset, we compared the performance of \mllm\ pretrained with and without this dataset. As shown in Table~\ref{tab:ab_nopre}, pretraining with FaceShield-pre10K improved fine-grained classification accuracy from $94.78\% \pm 0.19$ to $95.81\% \pm 0.11$, and reasoning accuracy from $98.83\% \pm 0.06$ to $99.29\% \pm 0.04$. Moreover, the Half Total Error Rate (HTER) was significantly reduced from $0.94\% \pm 0.05$ to $0.57\% \pm 0.04$, highlighting the dataset's effectiveness in improving both classification and interpretative capabilities.

The influence of data filtering and augmentation during supervised fine-tuning was analyzed using three configurations, as summarized in Table~\ref{tab:ab_filt_aug}. Without any filtering or augmentation, the model achieved a fine-grained classification accuracy of $87.92\% \pm 0.16$, reasoning accuracy of $98.14\% \pm 0.06$, and an HTER of $1.63\% \pm 0.05$. Adding data filtering improved these metrics to $90.88\% \pm 0.19$, $98.73\% \pm 0.03$, and $1.24\% \pm 0.08$, respectively. When both filtering and augmentation were applied, the model achieved the best performance with fine-grained classification accuracy of $95.81\% \pm 0.11$, reasoning accuracy of $99.29\% \pm 0.04$, and an HTER reduced to $0.57\% \pm 0.04$. These results underscore the complementary effects of filtering, which ensures task-relevant and semantically consistent data, and augmentation, which introduces linguistic variability and improves generalization.

\begin{table}[H]
\caption{Ablation Study on Pretraining w/ or w/o FaceShield-pre10K Dataset.}
\centering
\resizebox{\columnwidth}{!}{%
\begin{tabular}{cccccc}
\toprule[1pt]
\multicolumn{1}{c}{\parbox{2cm}{\centering \datasetpre}} &
\multicolumn{1}{c}{\parbox{4cm}{\centering Fine-grained \\ Classification}} &
\multicolumn{2}{c}{Reasoning} \\ 
\cline{3-4}
& \multicolumn{1}{c}{ACC (\%) $\uparrow$} & ACC (\%) $\uparrow$ & HTER (\%) $\downarrow$ \\ \hline
$\times$ & 94.78 $\pm$ 0.19 & 98.83 $\pm$ 0.06 & 0.94 $\pm$ 0.05 \\ 
\checkmark & \textbf{95.81 $\pm$ 0.11} & \textbf{99.29 $\pm$ 0.04} & \textbf{0.57 $\pm$ 0.04} \\ 
\bottomrule[1pt]
\end{tabular}%
}
\vspace{-0.6em}
\label{tab:ab_nopre}
\end{table}

\begin{table}[H]
\caption{Ablation Study on Data filtering and Data augmentation.}
\centering
\resizebox{\columnwidth}{!}{%
\begin{tabular}{ccccc}
\toprule[1pt]
\multicolumn{1}{c}{\parbox{2cm}{\centering Data \\ filtering}} & 
\multicolumn{1}{c}{\parbox{2cm}{\centering Data \\ augmentation}} &
\multicolumn{1}{c}{Fine-grained classification} & 
\multicolumn{2}{c}{Reasoning} \\ 
\cline{3-5}
 & & ACC(\%) $\uparrow$ & ACC(\%) $\uparrow$ & HTER(\%) $\downarrow$ \\ \hline
$\times$ & $\times$  & 87.92 $\pm$ 0.16 & 98.14 $\pm$ 0.06 & 1.63 $\pm$ 0.05  \\
\checkmark & $\times$  & 90.88 $\pm$ 0.19 & 98.73 $\pm$ 0.03 & 1.24 $\pm$ 0.08  \\
\checkmark & \checkmark & \textbf{95.81 $\pm$ 0.11} & \textbf{99.29 $\pm$ 0.04} & \textbf{0.57 $\pm$ 0.04}  \\ 
\bottomrule[1pt]
\end{tabular}%
}
\vspace{-0.6em}
\label{tab:ab_filt_aug}
\end{table}

Overall, the ablation study demonstrates the critical role of pretraining on FaceShield-pre10K, which provides foundational knowledge for downstream tasks, and the combined use of data filtering and augmentation, which enhances both data quality and linguistic diversity. By isolating these components, we confirm the robustness and effectiveness of the proposed training pipeline for multimodal anti-spoofing tasks.

\section{Evaluation Metrics}

To comprehensively evaluate the performance of the proposed model across different tasks, this study employs a range of evaluation metrics, including Half Total Error Rate (HTER)~\cite{yu2022deep}, Accuracy (ACC), BLEU~\cite{papineni2002bleu}, ROUGE-L~\cite{lin2004rouge}, METEOR~\cite{banerjee2005meteor}, and Average Precision (AP@40 and AP@50).

The choice of these evaluation metrics is motivated by the specific nature of each task. ACC and HTER are commonly used in Face Anti-Spoofing (FAS) tasks, providing a direct measure of model performance in distinguishing between genuine and spoofed faces. For the reasoning task, which involves generating long textual content for analysis, metrics such as BLEU, ROUGE-L, and METEOR are suitable as they are widely used in Natural Language Processing (NLP) to evaluate the quality of generated text. Lastly, for the localization task, where the goal is to locate the spoof area, we include object detection metrics like AP@40 and AP@50, which are effective in measuring the accuracy of the predicted regions of interest in detecting spoofed areas.

The detailed definitions and calculations of these metrics are presented as follows:

\paragraph{Half Total Error Rate (HTER)}
HTER is a critical metric for binary classification tasks, especially in scenarios with imbalanced data. It is defined as:
\begin{equation}
    \text{HTER} = \frac{\text{FAR} + \text{FRR}}{2},
\end{equation}
where \(\text{FAR}\) (False Acceptance Rate) represents the proportion of negative samples that are incorrectly classified as positive, and \(\text{FRR}\) (False Rejection Rate) represents the proportion of positive samples that are incorrectly classified as negative.

\paragraph{Accuracy (ACC)}  
Accuracy evaluates the proportion of correctly classified samples among all samples. It is defined as:  
\begin{equation}  
    \text{ACC} = \frac{\text{TP} + \text{TN}}{\text{Total}},  
\end{equation}  
where \(\text{TP}\) and \(\text{TN}\) are the true positives and true negatives, respectively, and \(\text{Total}\) is the sum of all samples.

\paragraph{BLEU (Bilingual Evaluation Understudy)}
BLEU is a metric commonly used in natural language generation tasks to evaluate the n-gram overlap between generated and reference texts. In this study, we utilize BLEU-1, BLEU-2, BLEU-3, and BLEU-4, which correspond to the 1-gram, 2-gram, 3-gram, and 4-gram match rates, respectively. The general formula for BLEU is:
\begin{equation}
    \text{BLEU-N} = \text{BP} \cdot \exp\left(\sum_{n=1}^N w_n \log p_n\right),
\end{equation}
where \(N\) is the maximum n-gram length, \(p_n\) is the proportion of matched n-grams, \(w_n\) is the weight (usually equally distributed), and \(\text{BP}\) is the brevity penalty to adjust for short generated texts:
\begin{equation}
    \text{BP} = 
    \begin{cases} 
    1 & \text{if } c > r, \\
    e^{(1-r/c)} & \text{if } c \leq r,
    \end{cases}
\end{equation}
where \(c\) and \(r\) are the lengths of the generated text and the reference text, respectively.

\paragraph{ROUGE-L (Recall-Oriented Understudy for Gisting Evaluation)}
ROUGE-L is based on the longest common subsequence (LCS) and measures the similarity between the generated and reference texts in terms of sequential word matches. It is defined as:
\begin{equation}
    \text{ROUGE-L} = \frac{\text{LCS}(\text{gen}, \text{ref})}{\max(\text{len}(\text{gen}), \text{len}(\text{ref}))},
\end{equation}
where \(\text{LCS}\) is the length of the longest common subsequence between the generated text \(\text{gen}\) and the reference text \(\text{ref}\), and \(\text{len}(\cdot)\) denotes the length of the respective text.

\paragraph{METEOR (Metric for Evaluation of Translation with Explicit ORdering)}
METEOR evaluates the semantic similarity between the generated and reference texts by considering word matching, stemming, and synonymy. It is computed as:
\begin{equation}
    \text{METEOR} = F_{\text{mean}} \cdot (1 - \text{Penalty}),
\end{equation}
where \(F_{\text{mean}}\) is the harmonic mean of precision and recall, and \(\text{Penalty}\) is a factor penalizing repeated patterns.

\section{Memory Requirement}

As shown in Tab.~\ref{tab:modelspecs}, the FaceShield model, with 3.83B parameters and a size of 10.42 GB, allocates 7.14 GB of memory during inference. This model achieves a balance between parameter count, model size, and memory allocation, making it suitable for environments with limited resources while maintaining strong performance.

\vspace{-2mm}
\begin{table}[h]
\renewcommand\arraystretch{1.2}
\centering
\scalebox{0.86}{
    \begin{tabular}{cccc}
    \toprule
    \textbf{Model} & \textbf{Parameters} & \textbf{Model Size} & \textbf{Memory Allocated} \\
    \midrule
    FaceShield & 3.83B & 10.42 GB & 7.14 GB \\
    \bottomrule
    \end{tabular}
}
\caption{Model Specifications}
\label{tab:modelspecs}
\vspace{-4mm}
\end{table}

\section{Detailed Experiments Setting}
\subsection{Detailed Setting of Model Architectures}
As illustrated in Fig.~\ref{fig:network}(a) depicts the baseline model, similar to LLaVA, comprising key components such as the Vision Encoder, Projector, Tokenizer, and LLM. This model only accepts RGB modality images as input. However, the high visual similarity between genuine faces and presentation attacks (PAs) in the RGB domain presents a significant challenge for this approach. In Fig.~\ref{fig:network}(b), the input is extended to include auxiliary modality images, which corresponds to the proposed SAVP strategy. Specifically, the auxiliary modality images are generated by applying operations such as Gray, HOG, and LBP to the RGB images, thereby enhancing the model's ability to perceive attack cues. In Fig.~\ref{fig:network}(c), we introduce the final model, which incorporates the PVTM module. This module retains and masks visual tokens based on the similarity between visual tokens and textual tokens, thereby reducing interference and redundancy. For a detailed explanation of the PVTM module, refer to Fig.~\ref{fig:network}(c).

Overall, the framework consists of a Siglip~\cite{zhai2023sigmoid} encoder for input processing, Phi-3~\cite{abdin2024phi} for inference, and a 2-layer MLP projector (with GELU activation in each layer) for feature mapping and output. The SAVP module utilizes dual-stream input to capture richer deception cues, although it may introduce redundancy. The PVTM mechanism effectively filters out unnecessary tokens, allowing the model to focus on the most important cues, ultimately improving performance.

\subsection{Detailed Setting of Comparison Methods}
In this study, we compare our proposed method with several baseline models, including PatchNet~\cite{Wang_2022_CVPR}, CoOp~\cite{zhou2022learning}, IADG~\cite{zhou2023instance}, LLaVA~\cite{liu2023llava}, Qwen-VL~\cite{Qwen-VL}, Minigpt4~\cite{zhu2023minigpt}, and Bunny~\cite{he2024bunny}. All of these comparison methods were implemented using the official open-source code, adhering to the specific settings and configurations provided by the respective authors. For the finetuned MLLMs, the training process was conducted using the dataset we introduced in this work.

\section{More Examples and  of Experimental Results of FaceShield}

\begin{figure}[H]
    \centering
    \includegraphics[width=1\linewidth]{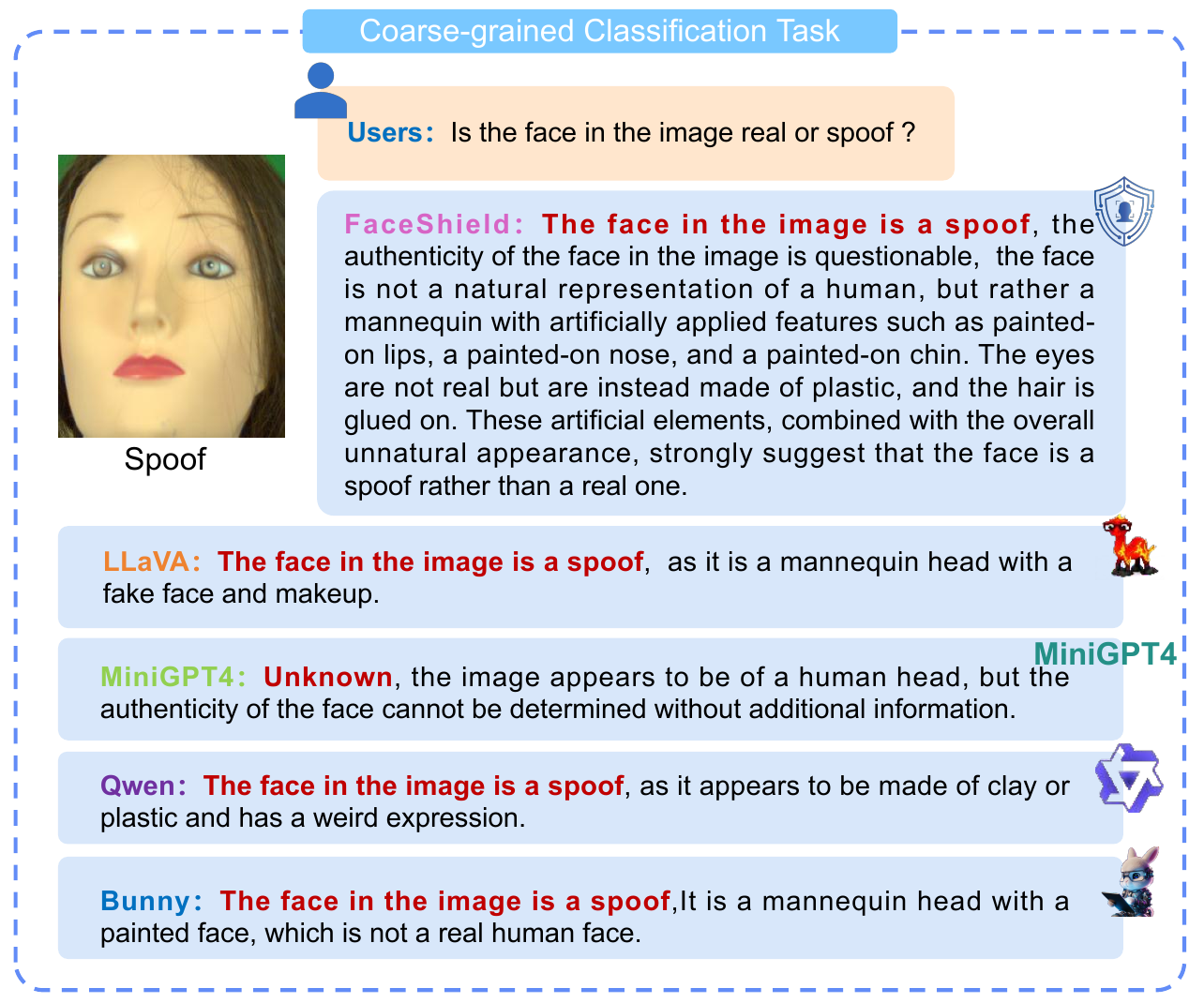}
    \caption{Example of Coarse-grained Classification Task}
    \label{fig:co_class_1}
\end{figure}

\begin{figure}[H]
    \centering
    \includegraphics[width=1\linewidth]{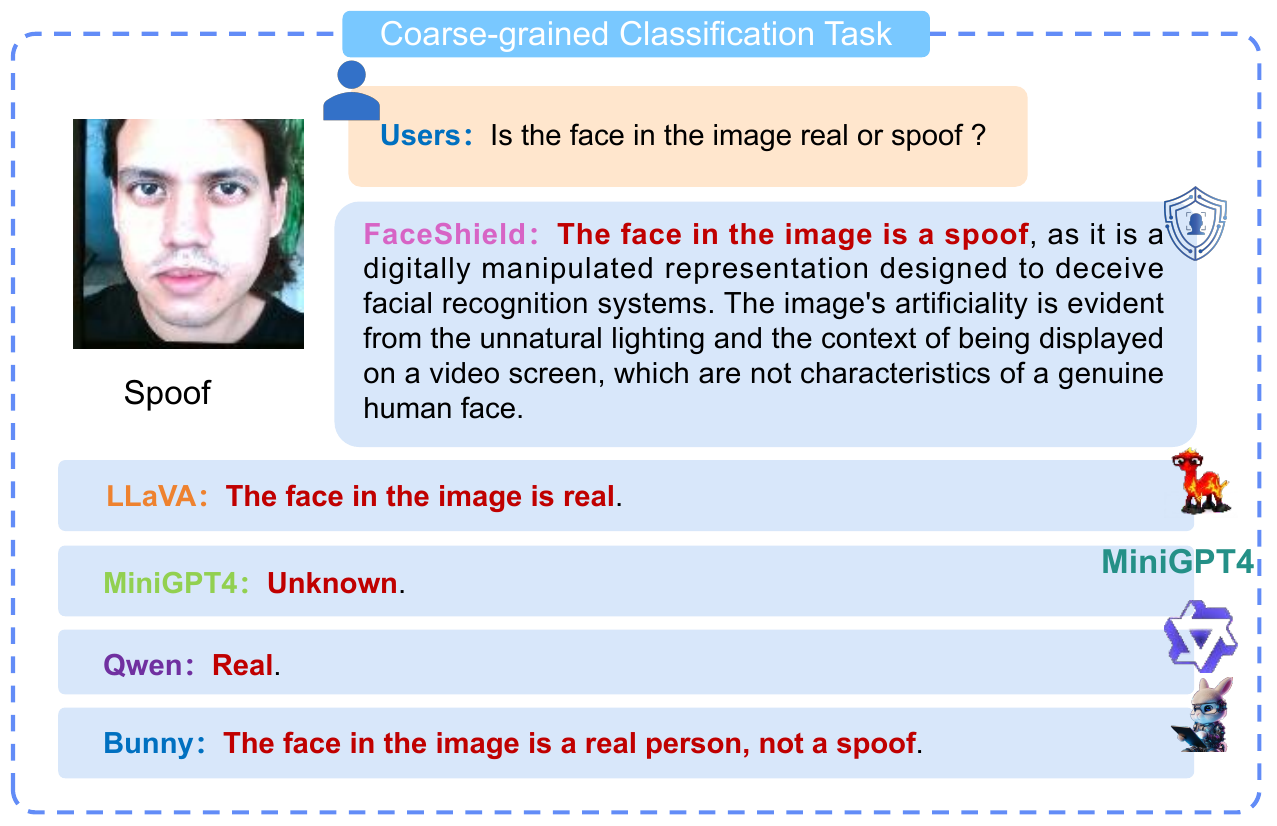}
    \caption{Example of Coarse-grained Classification Task}
    \label{fig:co_class_2}
\end{figure}

\begin{figure}[H]
    \centering
    \includegraphics[width=1\linewidth]{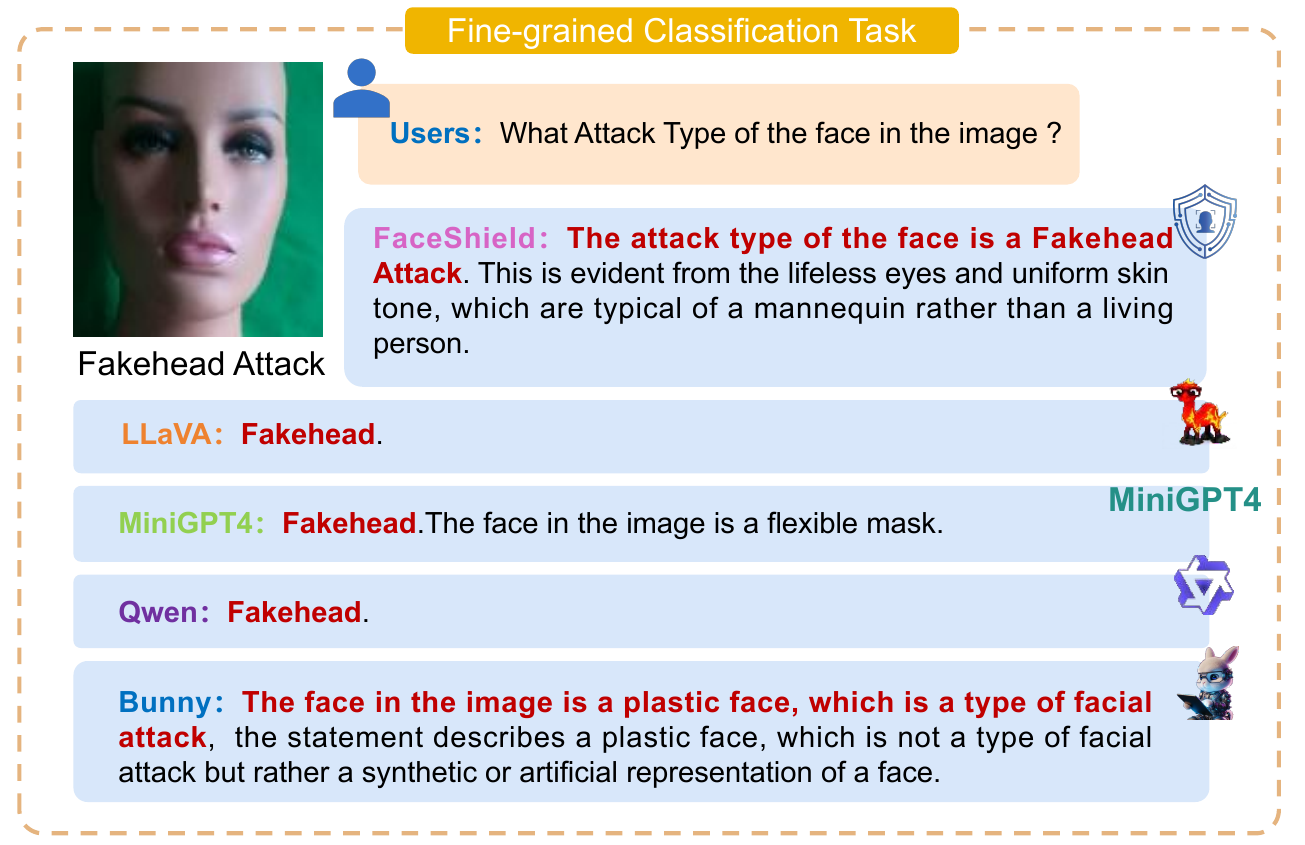}
    \caption{Example of Fine-grained Classification Task}
    \label{fig:fine_class_1}
\end{figure}

\begin{figure}[H]
    \centering
    \vspace{-1.3em}
    \includegraphics[width=1\linewidth]{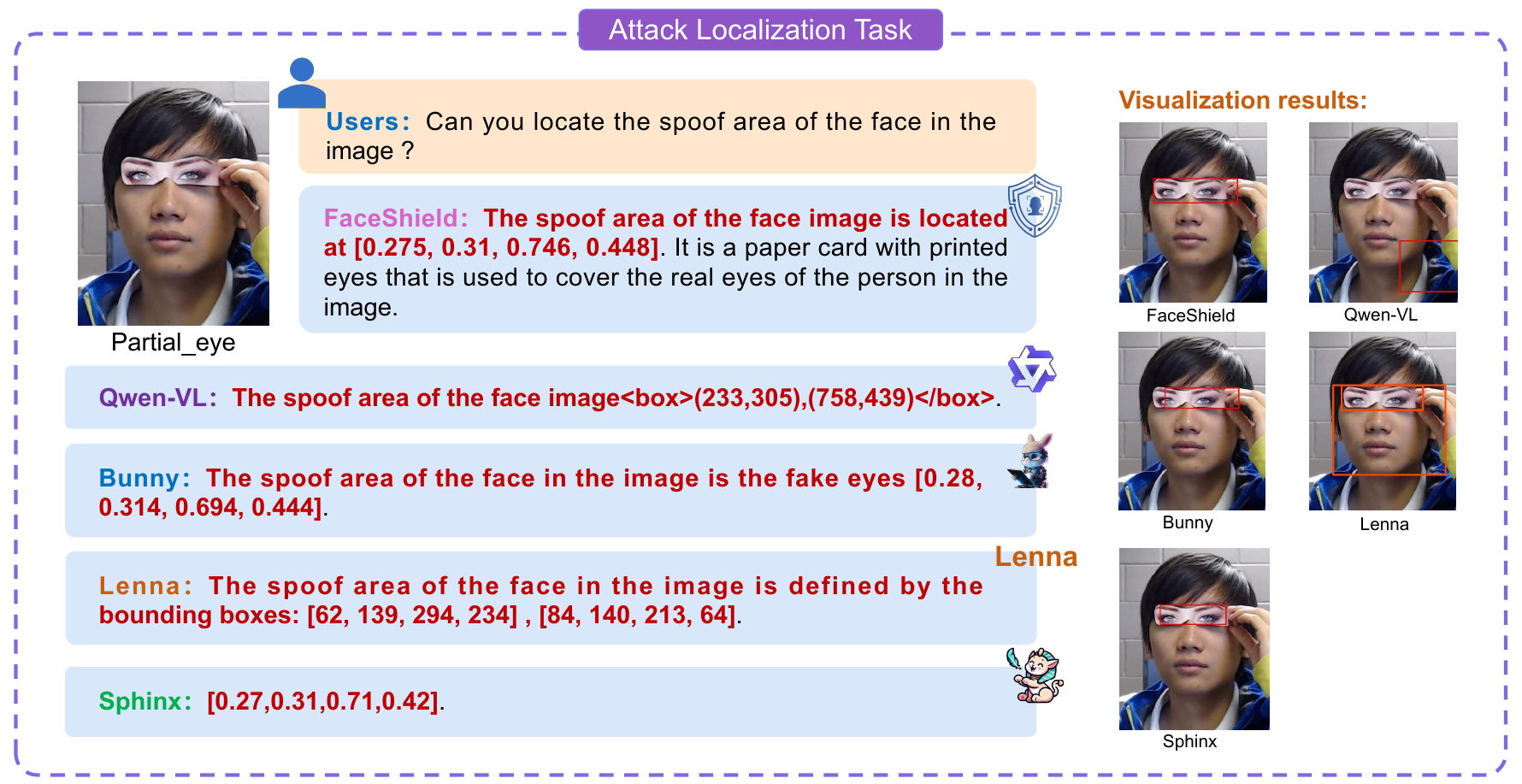}
     \vspace{-0.8em}
    \caption{Example of Attack Localization Task}
    \label{fig:localization}
\end{figure}

To illustrate the effectiveness of the proposed FaceShield model across various tasks, we present example outputs for the coarse-grained classification, fine-grained classification, reasoning, and attack localization tasks in Fig.~\ref{fig:co_class_1} to \ref{fig:localization}. These examples highlight the superior performance of FaceShield compared to other models, including LLaVA~\cite{liu2023llava}, Minigpt4~\cite{zhu2023minigpt}, QwenVL~\cite{Qwen-VL}, and Bunny~\cite{he2024bunny}.

\paragraph{Coarse-Grained Classification Task (Fig.~\ref{fig:co_class_1} and~\ref{fig:co_class_2})}
In Fig.~\ref{fig:co_class_1}, FaceShield correctly identifies the face as a spoof by observing critical features such as the unnatural gloss and artificial attributes of the face (e.g., it is a mannequin with painted features). Competing models, such as LLaVA and Qwen, provide inconsistent judgments, while MiniGPT-4 fails to determine the authenticity of the face. Similarly, in Figure~\ref{fig:co_class_2}, FaceShield recognizes the spoof face as being displayed on a screen, correctly leveraging cues such as unnatural lighting and the screen context. Other models, however, misclassify the face as real or fail to reach a conclusion. These results demonstrate that FaceShield excels in identifying subtle spoofing indicators, such as lighting anomalies and material inconsistencies, providing reliable binary classification for real and spoofed faces.

\paragraph{Fine-Grained Classification Task (Fig.~\ref{fig:fine_class_1})}
In Fig.~\ref{fig:fine_class_1}, FaceShield accurately classifies the spoof type as a ``Fakehead Attack,'' identifying key features such as lifeless eyes and uniform skin tone, which are typical of a mannequin. Competing models either fail to identify the correct spoof type or provide incomplete and inconsistent descriptions.

\begin{figure}[t]
    \centering
    \vspace{-1.3em}
    \includegraphics[width=0.95\linewidth]{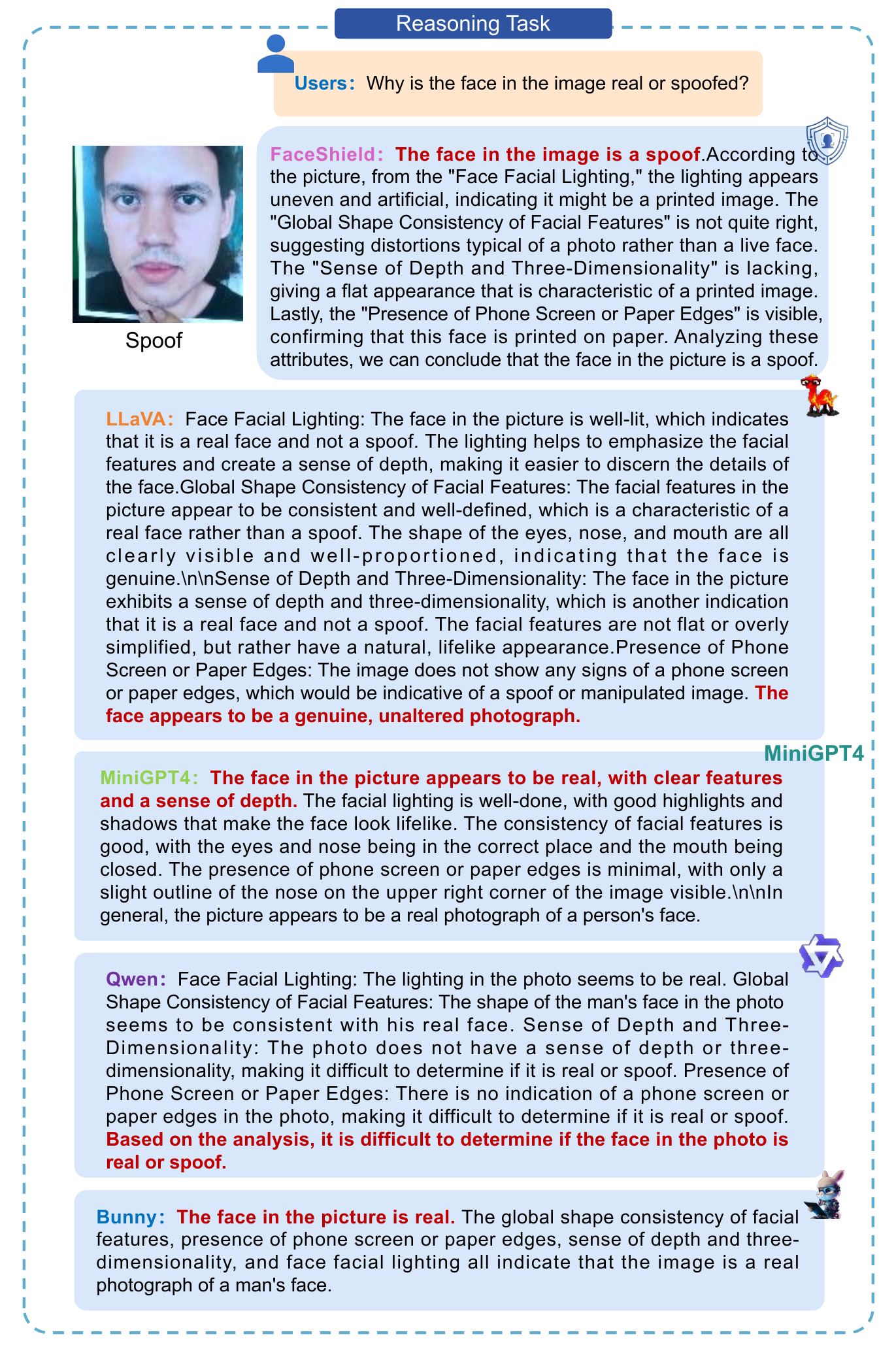}
     \vspace{-0.8em}
    \caption{Example of Reasoning Task}
    \label{fig:reasoning_1}
\end{figure}

\paragraph{Reasoning Task (Fig.~\ref{fig:reasoning_1})}
Fig.~\ref{fig:reasoning_1} demonstrate FaceShield's ability to provide comprehensive and interpretable reasoning for its decisions. In Figure~\ref{fig:reasoning_1}, FaceShield uses a structured reasoning framework based on four core attributes: facial lighting, global shape consistency, sense of depth and three-dimensionality, and the presence of edges indicative of spoofing. This structured approach allows FaceShield to conclude that the face is a spoof with high confidence. Competing models either misclassify the face as real or provide reasoning that lacks depth and critical observations.

\paragraph{Attack Localization Task (Fig.~\ref{fig:localization})}
In Fig.~\ref{fig:localization}, FaceShield precisely locates the spoofing region (e.g., the eyes covered by a paper card) and provides accurate bounding box coordinates. In contrast, other models provide vague or incorrect localization results. The precise identification of spoofing areas demonstrates FaceShield's strength in spatial reasoning and interpretability, which are critical for practical anti-spoofing applications.

\paragraph{Summary of Advantages}
Across all tasks, FaceShield consistently outperforms competing models due to its:
\begin{itemize}
    \item \textbf{Accurate Identification}: Superior coarse-grained and fine-grained classification of spoofing attacks with fewer errors.
    \item \textbf{Comprehensive Reasoning}: Structured and interpretable analysis based on core visual and contextual attributes.
    \item \textbf{Precise Localization}: Accurate and detailed identification of spoofing regions, enabling actionable insights.
\end{itemize}

These examples collectively highlight FaceShield's capability to address the multifaceted challenges in face anti-spoofing with a high degree of accuracy, interpretability, and robustness.

\begin{figure}[H]
    \centering
    \vspace{-0.9em}
    \includegraphics[width=1\linewidth]{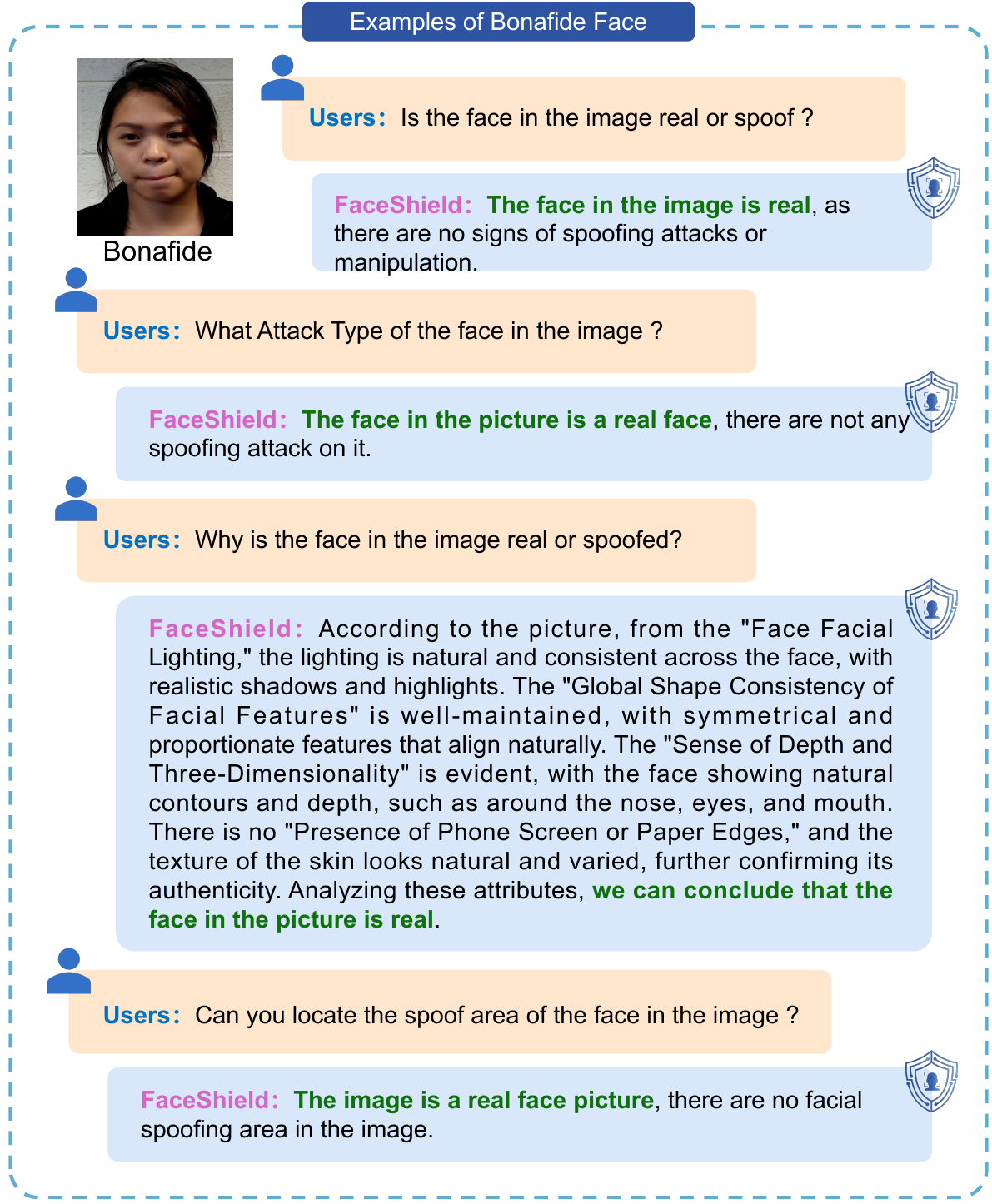}
    \caption{Example of Bonafide Face}
    \label{fig:bonafide_2}
\end{figure}

Furthermore, to highlight FaceShield’s accuracy in classifying on real human face images, we provide additional examples in Fig.~\ref{fig:bonafide_2}, further validating the model’s robust capability in handling genuine face data.

\begin{figure}[H]
    \centering
    \vspace{-0.9em}
    \includegraphics[width=1\linewidth]{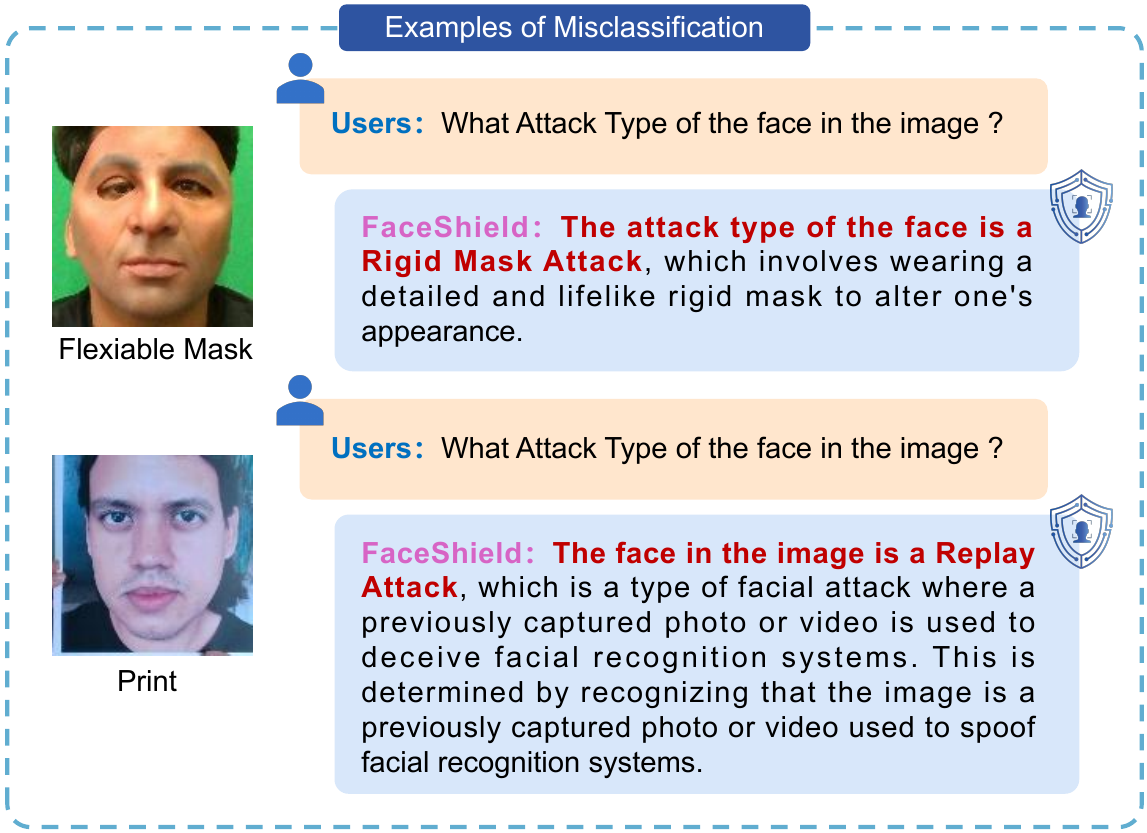}
    \caption{Example of Misclassification}
    \label{fig:misjudge}
\end{figure}

Additionally, Fig.~\ref{fig:misjudge} presents examples of misclassified cases. Analysis shows that frequent errors occur between Flexible and Rigid Masks, as well as Print and Replay attacks. These issues stem from insufficient data for Flexible and Rigid Masks, making it difficult to map material properties to attack types, while distinguishing Replay and Print attacks requires more contextual information.


\end{document}